\documentclass[10pt,twocolumn,letterpaper]{article}

\usepackage{cvpr}              % To produce the CAMERA-READY version
\usepackage{harpoon}% <---
\usepackage{esvect}
\usepackage{amsmath}
\usepackage{pifont}
\usepackage{multicol}
\usepackage{lipsum}
\usepackage[table]{xcolor}
\usepackage{gensymb}

\definecolor{cvprblue}{rgb}{0.21,0.49,0.74}
\usepackage[pagebackref,breaklinks,colorlinks,citecolor=cvprblue]{hyperref}

\def\paperID{8831} % *** Enter the Paper ID here
\def\confName{CVPR}
\def\confYear{2024}

\title{Edit One for All: Interactive Batch Image Editing}

\author{
Thao Nguyen \quad Utkarsh Ojha \quad Yuheng Li \quad Haotian Liu \quad Yong Jae Lee \\
University of Wisconsin--Madison\\
{\small \url{https://thaoshibe.github.io/edit-one-for-all}
}
}
\begin{document}

\makeatletter
\let\@oldmaketitle\@maketitle
\renewcommand{\@maketitle}{\@oldmaketitle
}
\maketitle
\makeatletter
% Image Editing has a long history in Computer Vision. However, prior work all focus on edit a single image at a time. In this paper, for the first time, we investigate batch image editing. Given an edit that already one in example photos, we propose a framework to transfer that edit into other test images. We divide edit types into two category: (1) Degree of the edit is fixed; or (2) Final state of the edit is fixed. We demonstrate that our framework can achieve better interactive batch image editing, which is save time and efforts in image editing process.
\begin{abstract}
\vspace{-5pt}
   % In the recent years, image editing has advanced remarkably. 
   % With increased human control, it is now possible to edit an image in a plethora of ways; from specifying in text what we want to change, to straight up dragging the contents of the image in an interactive point based manner. However, most of the focus has remained on editing single images at a time. How we can edit big batches of images has remained understudied. This paper presents a novel method for interactive batch image editing using StyleGAN as the medium. The goal is to reduce the need for human involvement \& effort so that a user is not needed in editing every image. Given an edit specified by users in an example image (e.g., make the face frontal), our method can automatically transfer that edit to other test images, so that regardless of their initial state (pose), they all arrive at the same final state (e.g., all facing front). Extensive experiments demonstrate that edits performed using our method have more visual consistency \& comparable quality to existing single-image-editing methods while saving significant time and human effort in the editing process.
In recent years, image editing has advanced remarkably. 
With increased human control, it is now possible to edit an image in a plethora of ways; from specifying in text what we want to change, to straight up dragging the contents of the image in an interactive point-based manner.
However, most of the focus has remained on editing single images at a time.
Whether and how we can simultaneously edit large batches of images has remained understudied.
With the goal of minimizing human supervision in the editing process, this paper presents a novel method for interactive batch image editing using StyleGAN as the medium.
Given an edit specified by users in an example image (e.g., make the face frontal), our method can automatically transfer that edit to other test images, so that regardless of their initial state (pose), they all arrive at the same final state (e.g., all facing front).
Extensive experiments demonstrate that edits performed using our method have similar visual quality to existing single-image-editing methods, while having more visual consistency and saving significant time and human effort.
\end{abstract}

%%%%%%%%%% 
\vspace{-5pt}
\section{Introduction}

% Image editing has undergone a transformation in recent years with the aid of modern generative models. The process has been greatly democratized, where many of the sophisticated edits, which previously might have required hours and niche expertise, can now be completed with relative ease in a matter of minutes. For example, different types of learning based algorithms can be used for image correction/adjustment~\cite{DeepPreset,PhotoExposure,MultipleExposure} \thao{Add more citations to color harmonization, exposure correction, and image blending} and for manipulating the semantic contents of real images ~\cite{StableDiffusion,InstructPix2Pix,SDEdit,DragDiffusion,DragGAN}. \thao{Add more citations to image editing methods}. In fact, there is no longer just \emph{one-way} to edit an image. The users can do so by specifying in text what change they want - e.g., \textit{make the hair darker} \cite{InstructPix2Pix,StyleCLIP,SDEdit}, or they can \textit{drag} the contents of the image in an interactive manner to shrink, enlarge or move a part of an image \cite{DragGAN,DragonDiffusion,UserControllableLT}.  

Image editing has undergone a transformation in recent years with the aid of modern generative models. The process has been greatly democratized, where many of the sophisticated edits, which previously might have required hours and niche expertise, can now be completed with relative ease in a matter of minutes. For example, different types of learning based algorithms can be used for image correction/adjustment~\cite{DeepPreset,PhotoExposure,MultipleExposure,DoveNet2020,tsai2017deep,Guo_2021_CVPR,ling2021region,jiang2021ssh,cong2022highresolution,valanarasu2022interactive,zhang2020deep,epstein2023selfguidance} and for manipulating the semantic contents of real images \cite{StableDiffusion,InstructPix2Pix,SDEdit,DragDiffusion,DragGAN,couairon2022diffedit,DragonDiffusion,yang2022paint,ControlNet,gligen}. Moreover, there are many different ways to edit an image.  For instance, a user can specify in text what changes they want - e.g., ``\textit{make the hair darker}'' \cite{InstructPix2Pix,StyleCLIP,SDEdit,kawar2023imagic}, or they can \textit{drag} the contents of the image in an interactive manner to shrink, enlarge, or move a part of an object \cite{DragGAN,DragonDiffusion,UserControllableLT,DragDiffusion,epstein2023selfguidance}.  

\begin{figure}[t]
\centering
\includegraphics[width=1\linewidth,page=1]{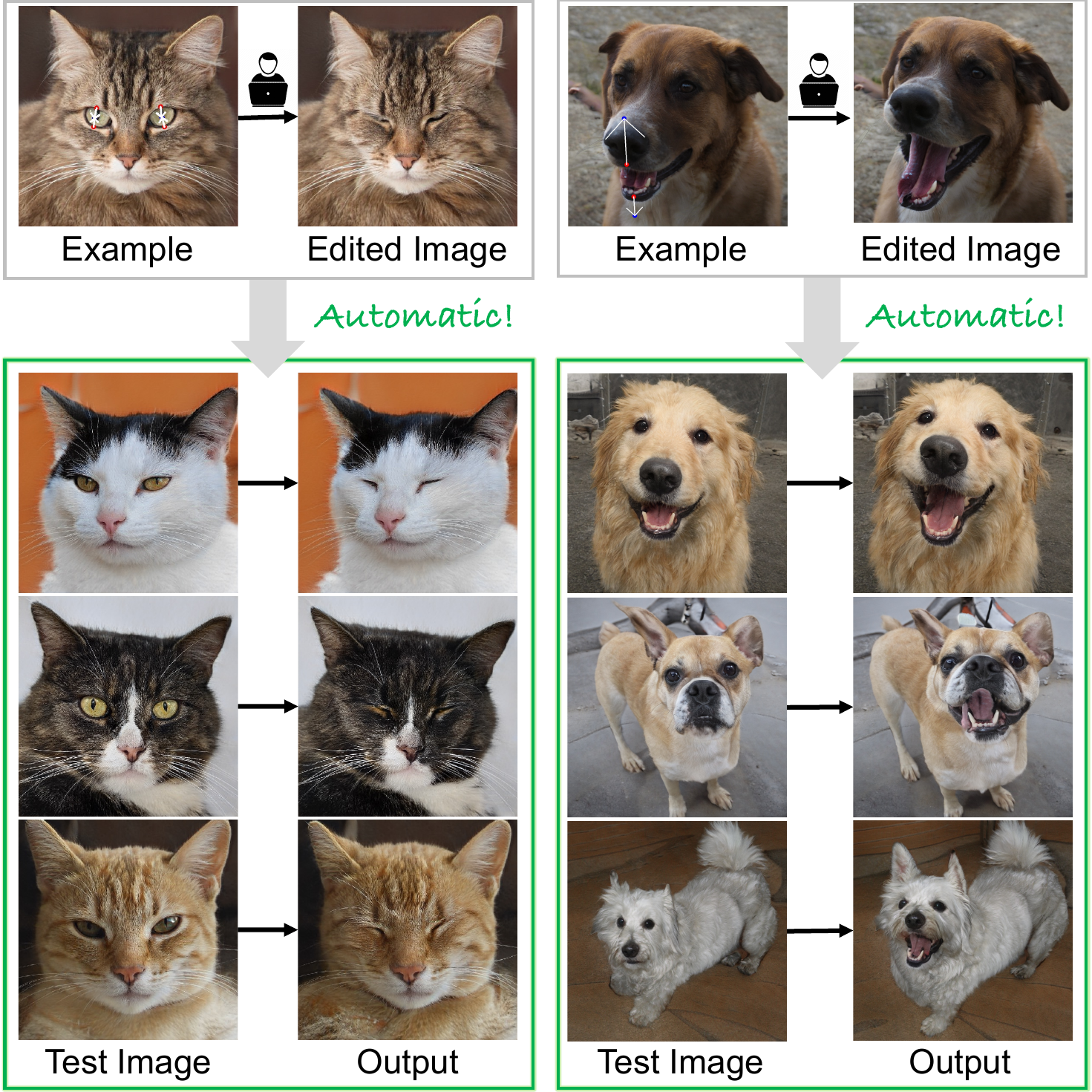}
\captionof{figure}{\textbf{Interactive Batch Image Editing.} Given a single user edited image, the goal is to automatically transfer that edit to new unseen images, so that all edited images end up with the same final state as the user edited image.}\label{fig:teaser}
\vspace{-3mm}
\end{figure}

\begin{figure*}[t!]
\centering 
\includegraphics[width=0.97\textwidth]{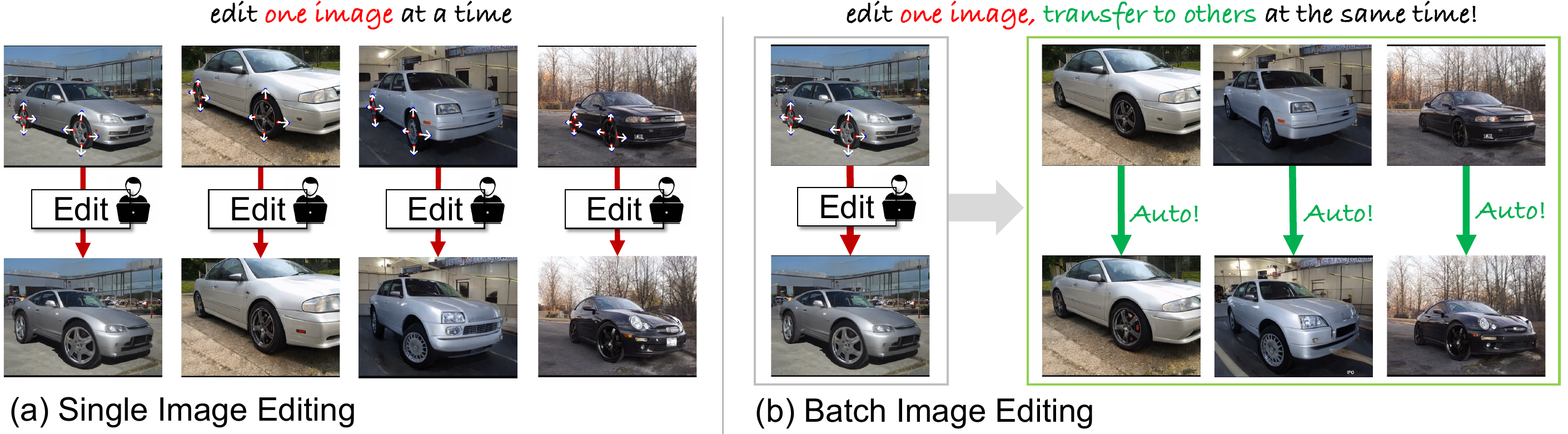}
% \fbox{\rule{0pt}{2in} \rule{.9\linewidth}{0pt}}
\vskip -0.1in
   \caption{\textbf{Single Image Editing vs. Batch Image Editing.} (a) Prior work (e.g., \cite{DragGAN,DragonDiffusion,UserControllableLT}) focuses on single image editing.  (b) We focus on batch image editing, where the user's edit on a single image is automatically transferred to new images, so that they all arrive at the same final state regardless of their initial starting state. In this way, we can achieve time speed up and reduce human effort in editing.}
   \vspace{-3mm}
\label{fig:batch-vs-single}
\end{figure*}

% A common theme across many of these works has been that the edits are typically designed to work for single images at a time. For example, given an image of a cat with open eyes, DragGAN~\cite{DragGAN} provides a way for us to drag the contents of \emph{that particular image} so that both the eyes can be closed in the resulting cat. But what if we wish to apply the same edit to \emph{many} different types of cats so that we can close all of their eyes? While one can have the user perform the dragging on each cat separately to make sure that happens, we can easily imagine that the whole process would be quite cumbersome (requiring lots of human annotation) and time intensive (requiring optimization for each image). In this paper, we study how one can automate this process. In particular, given a user-driven edit $x \rightarrow x'$, we propose a method to automatically transfer the edit from that example pair to other unseen images $\{y_1, y_2, ... y_n\}$ (same semantic category as $x$) so that all the edited images $\{y'_2, y'_2, ... y'_n\}$ have the same \emph{final state} that $x'$ had (e.g., all cats with eyes closed). We will follow the problem setup of \cite{DragGAN, StyleCLIP} and will use StyleGAN2~\cite{stylegan2} to perform the edits; i.e., an edit is made to an image $x \rightarrow x'$ by editing its representation in the StyleGAN's latent space, $w_x \rightarrow w_{x'}$.
%all the images ($x, x', y_1, y'_1 ...$) will be generated or, if real, will be inverted into its latent space for editing. Editing any image $x \rightarrow x'$

A common theme across many of these works is that the edits are designed to work for \emph{a single image} at a time. For example, given an image of a cat with open eyes, DragGAN~\cite{DragGAN} allows the user to drag the contents of \emph{that particular image} so that both the eyes can be closed in the resulting cat image. But what if we wish to apply the same edit to \emph{many} different types of cats so that we can close all of their eyes? While one could perform dragging on each cat separately, the whole process would be quite cumbersome (requiring lots of human annotation) and time intensive (requiring optimization for each image). 

In this paper, we introduce the new problem of \emph{Interactive Batch Image Editing}.  Given a user-specified edit $I \rightarrow I'$, the goal is to automatically transfer that edit to new unseen images, so that all edited images end up with the same \emph{final state} as $I'$ (e.g., all cats with eyes closed) regardless of their original starting states (e.g., varying degrees of original eye openness); see Figure~\ref{fig:teaser}. Hence, a solution to this problem requires two components: (i) modeling the user edit in the example pair $(I,I')$ so that it can be transferred to new images; and (ii) controlling the degree of the edit so that all edited images have the same final state.

To model the user's edit, we build upon existing literature on Generative Adversarial Networks (GANs)~\cite{härkönen2020ganspace,shen2020interpreting,shen2021closedform,StyleCLIP,LARGE}, and in particular StyleGAN~\cite{stylegan,stylegan2}, which shows that their learned latent space emits globally linear editing directions where the edited attribute (e.g., eye closeness) varies linearly in magnitude along such a direction for all semantically related instances (e.g., all closed eyes cat). To try to discover the global direction corresponding to a user edit, we perform optimization in the latent space of StyleGAN2~\cite{stylegan2} so that (i) the edited image with the discovered direction is visually similar to the original edited image $I'$, and (ii) the linear correlation between the distance along the direction and the strength of the visual attribute is highest (e.g., going twice the amount in that direction closes a cat's eyes by twice the amount).

To control the degree of the edit so that all edited images end up with the same final state, we derive a closed-form solution. Taking motivation from~\cite{LARGE}, we model the attribute strength (e.g., how much the cat's eyes are open) as the distance along the normal vector from a hyperplane. By setting the example edited image as lying on that hyperplane (i.e., its distance is 0), the objective is to move all other images along that same direction so that their distances to the hyperplane also become 0.  In this way, regardless of the start state of any image (e.g., varying degree of eye openness), their edited end states all become the same (e.g., closed eyes).  We show that we can analytically compute the exact amount of traversal along the direction for any image with a closed-form solution. Editing images in this way helps improve the results visually, as the final state of any new image matches that of the edited example given by the user.

Our method works with various edits given by an editing framework, e.g., it can be a change given by a dragging operation~\cite{DragGAN}, or it can be a change given by a text-based edit~\cite{StyleCLIP,InstructPix2Pix} (e.g., ``make eyes bigger''), as long as it is a direction compatible with a StyleGAN2~\cite{stylegan2} model.  We present qualitative and quantitative results in transferring edits for a variety of  objects (cats/dogs/faces/humans/lions/arts/etc.), parts (mouths/ears/legs/etc.), and corresponding attributes (big eyes/short faces/pale skins/etc.). Importantly, we show that the final state of the edited images is comparable to the scenario in which a user performs the edit for each image separately. But since our method does this automatically (see Fig.~\ref{fig:batch-vs-single}), we show that it takes much less time (e.g., only 0.05s per image, compared to DragGAN \cite{DragGAN} about 2s per image), and does not require laborious human annotation (e.g., only need 1 annotated image, compared to 4 required for DragGAN \cite{DragGAN}). Finally, we show practical applications of batch image editing; e.g., changing wheel size in every photo of a car collection by editing just one.

% Using the contribution of these two ideas, we present qualitative and quantitative effectiveness of our method in transferring edits to a batch of new images for a variety of domains; editing eyes of cat, mouths/ears of dogs, legs of horses, poses of human bodies etc. \lht{does this sound like our edit only works for some specific combinations? what about something like: a variety of target objects (cats/dogs/horses/humans); different parts (mouths, ears, legs, etc.)} Importantly, we show that the quality of our results, i.e, the final state of edited images, is comparable to the scenario in which a user is performing the edit for each image separately. But since our method does that automatically, we show that it takes much less time. Finally, we show some practical examples of batch image editing; e.g., changing your hair length in every photo of your wedding collection by editing just one. \lht{for some reason, when I see this example, I ask why? For such important events like the wedding, why not styling your hair to perfect at the beginning. (also asked my wife and she feels the same) But not sure.}

In sum, our contributions are: (1) We introduce the novel problem of interactive batch image editing, wherein a user-driven edit is automatically transferred to other similar test images. (2) We study what \emph{having the same final state} means in a geometrical sense, and propose a principled approach to achieving that in StyleGAN2~\cite{stylegan2}'s latent space. (3) We show that our method works on a wide variety of domains e.g., cats, dogs, humans etc., taking significantly less time than editing each image individually while having similar visual quality and more visual consistency.

\begin{figure*}[h]
\centering 
\includegraphics[width=1\textwidth,page=1]{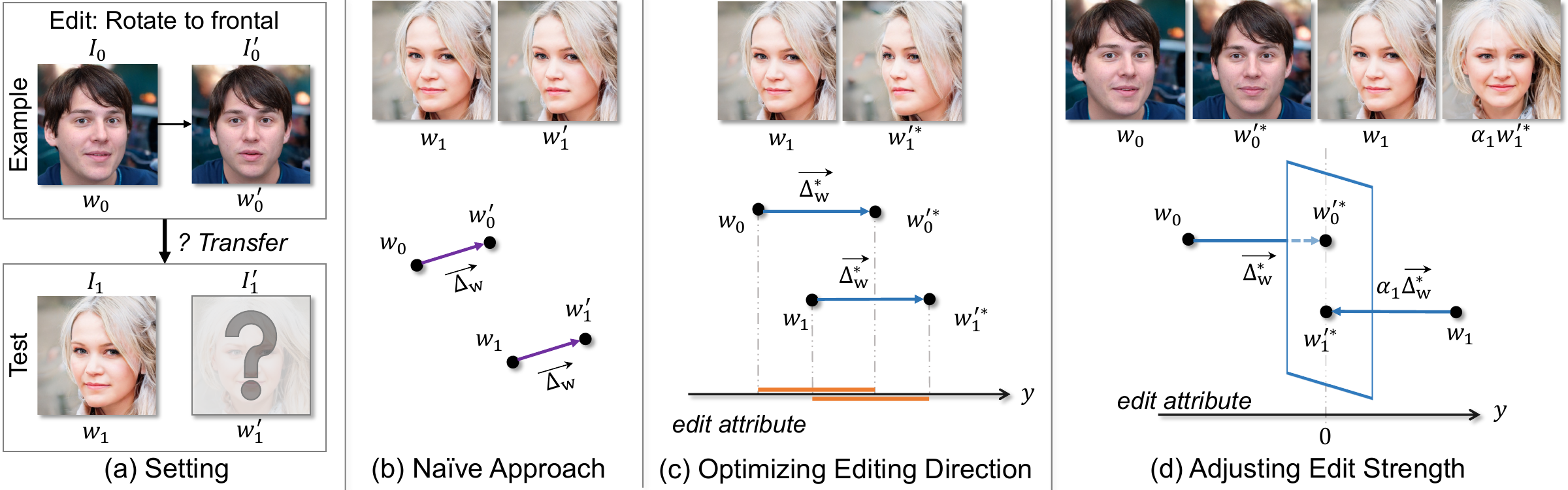}
% \fbox{\rule{0pt}{2in} \rule{.9\linewidth}{0pt}}
\vskip -0.1in
   % \caption{\textbf{Different editing strategies.} (a) The editing direction effective for an example image may not generalize well to test images. (b) Naive Approach: The editing direction effective for an example image may not generalize well to test images. (c) Optimizing Editing Direction: We optimize for a globally consistent direction that is effective for both example and test images. (d) Adjusting Editing Strength: Ensuring consistent final states requires adjusting the editing strength for each test image.}
   \caption{\textbf{Different editing strategies.} (a) Setting.
   (b) Naive Approach: The editing direction effective for an example may not generalize well to test images. (c) Optimizing Editing Direction: We optimize for a globally consistent direction that is effective for both example and test images. (d) Adjusting Editing Strength: Ensuring consistent final states requires adjusting the editing strength for each test image.}
   \vspace{-3mm}
\label{fig:illustration}
\end{figure*}

\section{Related Work}

\paragraph{Latent space of GANs.} The capabilities of generative adversarial networks (GANs) have transformed drastically, from the first GAN~\cite{goodfellow-nips2014} designed for simple image datasets like MNIST~\cite{lecun-mnist1998} to big and powerful models like BigGAN~\cite{brock2018biggan}, Progressive GAN~\cite{progan} and StyleGANs~\cite{stylegan, stylegan2} designed for much more complex datasets~\cite{imagenet, yu15lsun}. Parallel to the efforts to enhance their capabilities, there has also been work done to better understand and make use of their learned latent space. Specifically, there is a line of work which tries to find interpretable \emph{directions} in that space, so that moving in such a direction majorly changes only one discernible attribute in the image. Some methods \cite{gansteerability,bau2020units} try to find such directions or particular activations using (self) supervised methods. Others \cite{härkönen2020ganspace, shen2021closedform,shen2020interpreting} try getting rid of the need for supervision: discovering the directions representing the most important factors of variation, whatever they may be. Going beyond the $\mathcal{W}$ space, the authors in~\cite{stylespace} explore the $\mathcal{W+}$ space, and show that it has even better disentanglement useful for spatial image editing. All of these hidden capabilities have made StyleGANs a useful tool for editing purposes. However, prior work mainly focuses on the effectiveness of the editing direction (e.g., whether the discovered direction can change yaw pose). The question of whether these directions can be applied to diverse images to yield consistent results (e.g., all faces facing frontal) and, if so, how to achieving such consistency, remains unanswered.

\vspace{-10pt}
\paragraph{Image editing with generative models.} Researchers have used StyleGAN for \emph{real} image editing by designing an encoder to invert a real image into StyleGAN's latent space \cite{style_enc,Wang_2022_CVPR,e4e,pti}. StyleCLIP~\cite{StyleCLIP} presents a way to perform image editing through a text-based interface by making use of the CLIP encoder \cite{CLIP}. Very recently, StyleGAN has also been used to perform point-based editing \cite{UserControllableLT,DragGAN} so as to move any start point to reach a target location in the image; thereby elongating, rotating, shifting the objects. Most work on GANs has been for single object category datasets, but with the rise of text-to-image Diffusion Models which can generate complex images \cite{StableDiffusion,LatentDiffusion,DDPM}, editing such images is now possible. Complex scenes can be edited either using text \cite{mokady2022null,hertz2022prompt, InstructPix2Pix,kawar2023imagic,SDEdit,ControlNet}, or using the same idea of point-based manipulation \cite{DragDiffusion, DragonDiffusion}. There have been some works which discuss the possibility to transfer the edit of one image to another. EditGAN~\cite{EditGAN} uses segmentation mask manipulations to edit an image, but to successfully transfer them to an unseen image, it needs to do post hoc manipulation of the editing scale. RewriteGAN~\cite{bau2020rewriting} involves editing one generated sample; e.g., adding a patch of a tree onto a church tower. Following this, the rules of the GAN are manipulated so that all churches have \emph{some} tree on the top. Within diffusion models, visual prompting/ image analogies tries something similar, where users can define a triplet \{before, after, test\} to learn the edit and transfer it to a test image \cite{bar2022visual,visii,sun2023imagebrush,image_analogy,deep_image_analogy}. Our task is similar, in that we wish to transfer user-edits from the training example to new images, but it differs in one crucial aspect: along with transferring the edit, we wish to automatically learn its strength so that the edit produces the same final state for a new image.

\section{Approach}

Our focus is on the setting in which an image editing process edits an \textit{attribute} of an image so that the attribute's \emph{value} reaches a desired state; e.g., rotating the pose of a face looking sideways so that it faces front (attribute = face's pose, value = front). 

With this view, we explain our framework for \emph{batch image editing}, which can be broken down into two stages: (i) A user edits an image $I_{0}$ (e.g., using DragGAN~\cite{DragGAN}) to obtain an edited image $I'_{0}$. We describe in Sec.~\ref{sec:model_edit} how we capture this user edit $I_{0} \rightarrow I'_{0}$, so that the same edit can be applied to new images. (ii) Next, we describe in Sec.~\ref{sec:adjusting}, how for any new image, we apply the modeled edit by automatically adjusting its \emph{strength} so that the attribute's value in this new image matches that of $I'_{0}$ (e.g., any face, regardless of its initial pose, now faces front after the edit). 

% The goal of an image editing process is to edit an \textit{attribute} of an image so that the attribute's \emph{value} reaches a desired state for the human; e.g., rotating the pose of a face looking sideways so that it faces front (attribute = face's pose, value = front). With this view, we now explain our framework for batch image editing, which can be broken down into two stages: (i) A user edits an image $I_{0}$ using DragGAN~\cite{DragGAN} or StyleCLIP~\cite{StyleCLIP} to obtain an edited image $I'_{0}$ \thao{should we change to StyleGANs in general? Specify StyleCLIP and DragGAN here make me feel like narrow our scope.}. We describe in Sec.~\ref{sec:model_edit} how we capture this user edit $I_{0} \rightarrow I'_{0}$, so that the same edit can be applied to new images. (ii) Next, we describe how in Sec.~\ref{sec:adjusting} how for each test image, we apply the modeled edit by automatically adjusting its \emph{strength} so that the attribute's value in this test image matches that of the ground-truth edit (e.g., any face, regardless of its initial pose, now faces front after the edit).  

\subsection{Modeling the User Edit}\label{sec:model_edit}

We start with an image $I_{0}$. This image could be a real image or a generated one. Either way, we get its latent representation in the $\mathcal{W}$ space of a StyleGAN2~\cite{stylegan2} model $G$, so that $I_{0} = G(w_{0})$.  (For real images we can use GAN-inversion techniques~\cite{xia2022gan}.)  The user, with the help of an image editing framework (e.g., DragGAN~\cite{DragGAN} or InstructPix2Pix~\cite{InstructPix2Pix}), edits $I_{0}$ to manipulate one or more of its attributes. The edit maps to the $\mathcal{W}$ space as $w_{0} \rightarrow w'_{0}$. The resulting edited image can thus be recovered as follows: $I'_{0} = G(w'_{0})$. Fig.~\ref{fig:illustration}(a) shows an example of the original and edited image pair, ($I_{0}, I'_{0}$), where the user intended to turn the face forward.
% \yh{this may cause confusion: here it says image I, and asks readers to check figure 3, but there is no notation I in the figure}

%We start with an image $I_{0}$. This image could be a real image or a generated one. Either way, we get its latent representation in the $\mathbf{w}$ space of a StyleGAN2~\cite{stylegan2} model $G$, so that $I_{0} = G(w_{0})$.  The user, with the help of an image editing framework - e.g., DragGAN~\cite{DragGAN} or StyleCLIP~\cite{StyleCLIP}, edits $I_{0}$ to manipulate one or more of its attributes. The edit takes place in the $w$ space: $w_{0} \rightarrow w'_{0}$. The resulting image can thus be obtained as follows: $I'_{0} = G(w'_{0})$. Fig.~\ref{fig:illustration}(a) shows an example of the training pair, ($I_{0}/I'_{0}$), where the user intended to turn the face forward.

Now, given the user edit $I_{0} \rightarrow I'_{0}$, we wish to capture it in a way that can be applied to new images in a generalizable manner; i.e., the application of the edit changes the \emph{same} property in a new image. This is where a nice property of GANs, and in particular StyleGANs~\cite{stylegan, stylegan2} comes in useful. It has been shown in prior works~\cite{härkönen2020ganspace, shen2020interpreting, shen2021closedform} that it is possible to discover directions with such properties (using supervised as well as unsupervised methods) in the learned $\mathcal{W}$ space. In particular, it is possible to find directions ($\Delta_{w}$) that are \emph{globally consistent}. Taking motivation from \cite{LARGE}, we define a globally consistent direction $\Delta_{w}$ as the following: for any arbitrary $w$, moving along $\Delta_{w}$, $w \rightarrow w + \Delta_{w}$ (i) changes the same attribute, and (ii) by the same amount. 

% Now, given the user edit $I_{0} \rightarrow I'_{0}$, we wish to capture it in a way that can be applied to new images in a generalizable manner; i.e., the application of the edit changes the \emph{same} property in a new image. This is where a nice property of GANs, and in particular StyleGANs~\cite{stylegan, stylegan2} comes in useful. It has been shown in prior works~\cite{härkönen2020ganspace, shen2020interpreting, shen2021closedform} that it is possible to discover edits with such properties (using supervised as well as unsupervised methods) in the learnt $\mathbf{w}$ \thao{$\mathcal{W}$?} space. In particular, it is possible to find directions ($\Delta_{w}$) in that space that are \emph{globally consistent}. Taking motivation from \cite{LARGE}, we define a globally consistent direction $\Delta_{w}$ as the following: for any arbitrary $w$, moving along $\Delta_{w}$, $w \rightarrow w + \Delta_{w}$ (i) changes the same attribute, and (ii) by the same amount. 

To make the precise user edit $I_{0} \rightarrow I'_{0}$ applicable to other images, it needs to be captured as a globally consistent direction. The naive way to represent that edit will be through a simple difference in the $\mathcal{W}$ space: $\Delta_{w} = w'_{0} - w_{0}$. However, empirically, and as we show in Fig.~\ref{fig:illustration}(b), applying this $\Delta_{w}$ to the latent code $w_{1}$ corresponding to a new image $I_{1}$ \emph{does not} always result in the same change; while $I_{0} \rightarrow I'_{0}$ results in a pose change of $\sim$30\degree degrees in yaw, $I_{1} \rightarrow I'_{1}$ does not seem to change the same attribute or at least not by the same amount. 

%To make the (very particular) user edit $I_{0} \rightarrow I'_{0}$ applicable to other images, it needs to be captured as a globally consistent direction. The naive way to represent that edit will be through a simple difference in the $\mathbf{w}$ \thao{\thao{$\mathcal{W}$?}} space: $\Delta_{w} = w'_{0} - w_{0}$. However, as we show in Fig.~\ref{fig:illustration}(b), applying this $\Delta_{w}$ to the latent code $w_{1}$ corresponding to a new image $I_{1}$ \emph{does not} result in the same change; while $I_{0} \rightarrow I'_{0}$ resulted in a pose change, $I_{1} \rightarrow I'_{1}$ does not seem to change the same attribute. \thao{I'm not sure about it. Maybe the images I choose is not good. But I think: $I_{1} \rightarrow I'_{1}$ change the same attribute, just not the desired amount (e.g., to frontal)}

Hence, our goal is to represent the $I_{0} \rightarrow I'_{0}$ edit through a $\mathcal{W}$ space direction that can better satisfy both the properties of a globally consistent direction. For this task, we first take motivation from LARGE~\cite{LARGE} to introduce a mathematical view of what those directions mean. 

% Hence, our goal is to represent the $I_{0} \rightarrow I'_{0}$ edit through a $\mathbf{w}$ space direction that can satisfy both the properties of a globally consistent direction. For this task, we first take motivation from LARGE~\cite{LARGE} to introduce a mathematical view of what those directions mean. 

\vspace{-10pt}
\paragraph{Globally consistent direction.} Let's say that $\Delta_{g}$ is one such direction. Along with a bias term $b$, we can define a hyperplane as follows:
\begin{equation}
     {w} \cdot {\Delta_{g}}  + b = 0
    \label{eq:hyperplane}
\end{equation}
That is, any point $w$ which lies on the hyperplane will satisfy this equation. The authors in~\cite{LARGE} argued that for such a hyperplane, whose normal vector is a globally consistent direction ($\Delta_{g}$), the distance of an arbitrary point $w$ from that hyperplane ($w \cdot {\Delta_g}$) will be linearly correlated to the actual attribute ($y$) that results in the generated image:
\begin{equation}
    y = a \times ({w} \cdot {\Delta_g}) + b
    \label{eq:linear_correlation}
\end{equation}
Here, $a$ and $b$ are unknown linear coefficients. For example, if $\Delta_{g}$ corresponds to \textit{change in pose}, then all the front facing people will lie at the same distance $d$ from the above hyperplane. Because of this, any hyperplane defined with respect to a globally consistent direction can be viewed as a semantic hyperplane. For simplicity, we can set $d=0$ for the edited image $I'_{0}$. 

% \paragraph{Globally consistent direction:} Let's say that $\Delta_{g}$ is one such direction. Along with a bias term $b$, we can define a hyperplane as follows:
% \begin{equation}
%     {\Delta_{g}} \cdot {w} + b = 0
%     \label{eq:hyperplane}
% \end{equation}
% That is, any point $w$ which lies on the plane will satisfy this equation. The authors in ~\cite{LARGE} argued that for such a hyperplane, whose normal vector is a globally consistent direction ($\Delta_{g}$), the distance of an arbitrary point $w$ from that hyperplane ($w \cdot {\Delta_g}$) will be linearly correlated to the actual attribute ($y$) that results in the generated image.

% \vspace{-0.3cm}
% \begin{equation}
%     y = a \times ({w} \cdot {\Delta_g}) + b
%     \label{eq:linear_correlation}
% \end{equation}
% Here, $a$ and $b$ are unknown linear coefficients. For example, if $\Delta_{g}$ corresponds to \textit{change in pose}, then all the front facing people will lie at some same distance $d$ from the above hyperplane. Because of this, any hyperplane defined with respect to a globally consistent direction can be viewed as a semantic hyperplane. For simplicity in our experiments, we set $d=0$ for the edited image $I'_{0}$. 

Given that the original $\Delta_{w}$ may not always be globally consistent, we aim to discover, through optimization, a different direction $\Delta^{*}_{w}$ which produces a similar editing effect, but is more likely to be a globally consistent one. We initialize $\Delta^{*}_{w}$ with 0's, and design two objective functions to optimize it. First, to make sure that it produces a similar effect as the original direction, we use an image reconstruction loss so that the edited image produced by the new edit matches the original edit given by the user.
% \thao{We choose to optimize in the image space rather than the latent space for better robustness.}
\begin{equation}
    \mathcal{L}_{img} = \left\| G({w_0} + {\Delta_{w}}) - G({w_0} + {\Delta^{*}_{w}})\right\|_{2}
\end{equation}

% Additionally, since we have set the distance of the edited image from the hypothetical hyperplane to be 0, we can constrain the learned $\Delta^{*}_{w}$ to follow this property.
Additionally, user can provide the real value of distance $d$, otherwise we set the distance of the edited image from the hypothetical hyperplane to be 0 (similar to \cite{LARGE}). We constrain the learned $\Delta^{*}_{w}$ to follow this property for better interpretability.
That is, the new direction should be such that when we traverse the original latent code in that direction, $w_{0} \rightarrow w_{0} + \Delta^{*}_{w}$, the resulting image should lie at the hyperplane defined by that direction. Setting $b=0$ in Eq.~\ref{eq:hyperplane}, we minimize the edited point's distance from the hyperplane:
\begin{equation}
    \mathcal{L}_{att} = \left| (w_{0} + \Delta^{*}_{w}) \cdot \Delta^{*}_{w}  \right|
\end{equation}

% Additionally, since we have set the distance of the edited image from the hypothetical hyperplane to be 0, we constrain the learnt $\Delta^{*}_{w}$ to follow this property. That is, the new direction should be such that when move the original latent code in that direction, $w_{0} \rightarrow w_{0} + \Delta^{*}_{w}$, the resulting image should lie at the hyperplane defined by that direction. Setting $b=0$ in Eq.~\ref{eq:hyperplane}, we minimize the edited point's distance from the hyperplane:
% \begin{equation}
%     \mathcal{L}_{att} = \left| (w_{0} + \Delta^{*}_{w}) \cdot \Delta^{*}_{w}  \right|
% \end{equation}

The overall loss function for optimizing the new direction $\Delta^{*}_{w}$ is a weighted sum of the two objectives: $\mathcal{L}_{\Delta} = \mathcal{L}_{img} + \lambda \mathcal{L}_{att}$. We set $\lambda$ to 0.02 in all experiments to balance the magnitude of $\mathcal{L}_{img}$ and $\mathcal{L}_{att}$.
% Please refer to the supplementary for the pseudo-code. 

%The overall loss function for training the new direction $\Delta^{*}_{w}$ is a weighted sum of the two objectives: $\mathcal{L}_{\Delta} = \mathcal{L}_{img} + \lambda \mathcal{L}_{att}$. We set $\lambda$ to 0.02 in all experiments to balance the magnitude of $\mathcal{L}_{img}$ and $\mathcal{L}_{att}$. Please refer to the Supplementary for the pseudo-code. 

\begin{figure*}[t]
\centering 
\includegraphics[width=\textwidth]{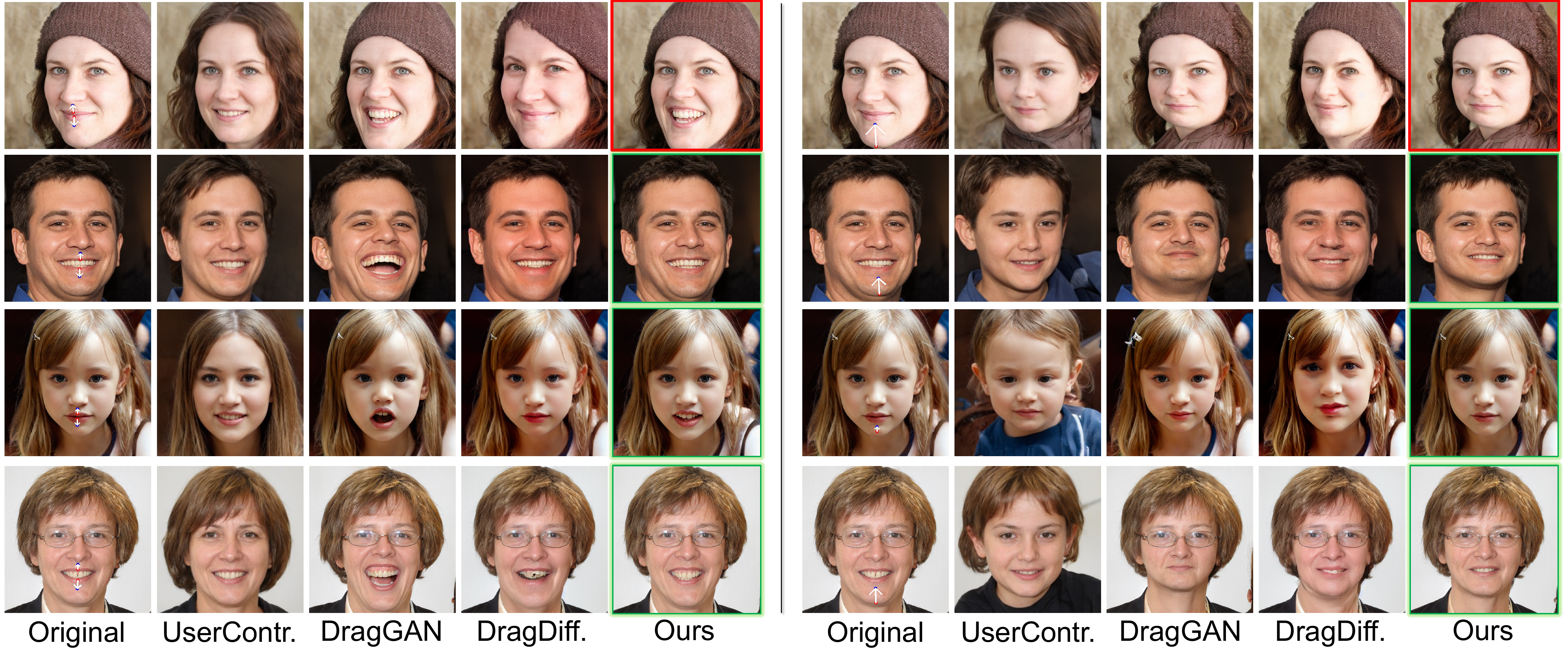}
% \fbox{\rule{0pt}{5in} \rule{.9\linewidth}{0pt}}
\vskip -0.1in
   \caption{Qualitative comparisons between dragging baselines. For ours, green bounding box indicates automatic transfer from the red bounding box example in the first row (i.e., no point annotation needed!).}
   \vspace{-3mm}
\label{fig:comparison-drag}
\end{figure*}
%  Our edited images are generally more consistent.

Now, in Fig.~\ref{fig:illustration} (b) and (c), we compare the difference in image editing results using two different directions when applied to the same new image ($I_{1}$): $w_{1} \rightarrow w_{1} + \Delta_{w}$ vs. $w_{1} \rightarrow w_{1} + \Delta^{*}_{w}$. As discussed before, in this case, the editing using the naively computed direction ($\Delta_{w}$) is not able to accurately capture the pose rotation the way it did for the user edited example ($I_{0} \rightarrow I'_{0}$). On the other hand, with $\Delta^{*}_{w}$, we see that the pose of the woman changes by a similar amount as the edited example. However, since the original pose of the woman ($I_{1}$) was not the same as the pose of the man ($I_{0}$), the new edited image ($I'_{1}$) still does not arrive at the same final state as $I'_{0}$. Therefore, our next goal is to figure out how to scale the learned direction so that $I'_{1}$ does arrive at the same final state as $I'_{0}$.

\subsection{Adjusting Editing Strength for New Images}
\label{sec:adjusting}

Given the optimized $\Delta^{*}_{w}$ and some new image $I_{i} = G(w_{i})$, we wish to edit it in the following way:
\begin{equation}
    w'_{i} = w_{i} + \alpha_{i}n
\end{equation}
where $\alpha_{i}$ is the editing strength computed separately for each image and $n = \Delta^{*}_{w} / || \Delta^{*}_{w} ||$ is a unit vector in the direction of $\Delta^{*}_{w}$. Fig.~\ref{fig:illustration} (d) illustrates a geometric perspective that we will use to compute $\alpha_{i}$. First, we see $\Delta^{*}_{w}$ represented as a vector and its corresponding hyperplane that is normal to it. Next, we depict the new latent point $w_{i}$. The goal is to move it along the $\Delta^{*}_{w}$ direction so that it arrives \emph{onto} the hyperplane. Through this depiction, $\alpha_{i}$ can be understood as the distance of $w_{1}$ from the hyperplane. We can get the distance by projecting ($w'_{0} - w_{i}$) in the direction of $n$. Therefore, $\alpha_i = (w'_{0} - w_{i}) \cdot n$.

% Given the learnt $\Delta^{*}_{w}$ and some test image $I_{i} = G(w_{i})$, we wish to edit it in the following way:
% \begin{equation}
%     w'_{1} = w_{1} + \alpha_{i}n
% \end{equation}
% where $\alpha_{i}$ is the editing strength computed separately for each image and $n = \Delta^{*}_{w} / || \Delta^{*}_{w} ||$ is a unit vector    . Fig.~\ref{fig:illustration}(c) illustrates a geometric perspective that we will use to compute $\alpha_{i}$. First, we see $\Delta^{*}_{w}$ represented as a vector (\ut{Thao: make sure to make the adjustments}) and its corresponding hyperplane that is normal to it. Next, we depict the test latent point $w_{1}$. The goal is to move it along the $\Delta^{*}_{w}$ direction so that it arrives \emph{onto} the hyperplane. Through this depiction, $\alpha_{i}$ can be understood as the distance of $w_{1}$ from the hyperplane. We can get the distance by projecting ($w'_{0} - w_{1}$) in the direction of $n$. Therefore, $\alpha_i = (w'_{0} - w_{1}) \cdot n$.

Since for each image, the only unique computation that needs to be done is the calculation of $\alpha_{i}$, we can see why our method will be much faster than, for example, annotating every image and running the DragGAN~\cite{DragGAN} optimization each of those times. We will show in our experiments the significant difference in time taken by our method compared to single image editing baselines. 
%\thao{Can we add here: Also if running DragGAN on each image, you will also annotate each images. This is significantly more laborious that ours.}

Importantly, there is an added benefit to computing $\alpha$'s in this way. Let's say the goal of the editing process is to rotate $n$ faces and make them frontal. After completion, all of them do become frontal, each with their own editing strengths $\{\alpha_{1}, \alpha_{2}, ..., \alpha_{n}\}$ computed using the above formula. Suppose the user now wants the same faces facing a bit left instead. To do this, the user \emph{does not} need to re-annotate the original training example and run the optimization one more time. In an interactive manner, they can simply scale the $\alpha$ for the training example to match the desired edit, and all other $\alpha$'s can be automatically recomputed.

\begin{figure*}[t]
\centering 
\includegraphics[width=1\textwidth]{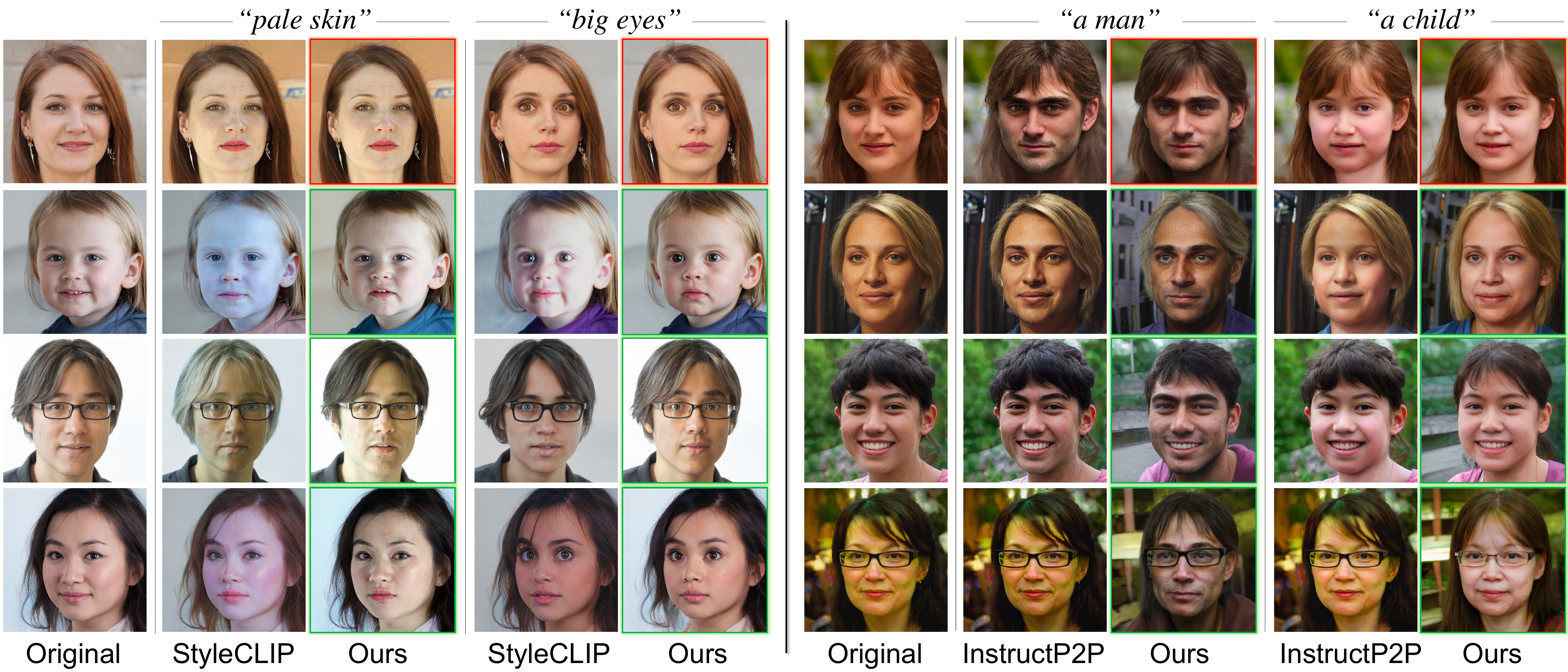}
% \fbox{\rule{0pt}{5in} \rule{.9\linewidth}{0pt}}
\vskip -0.1in
   \caption{Qualitative comparisons to text-guided baselines. Ours transfers the edit from example (red), to other test images (green).}
   \vspace{-3mm}
\label{fig:comparison-text}
\end{figure*}

\section{Experiments}

We study how well our method models and transfers the edits from an example, and how efficient it is compared to single image editing baselines using their official code.% In Sec.~\ref{sec:qual}, we first show the qualitative results visualizing the consistency of the transferred edits using different methods. Following that, in Sec.~\ref{sec:quant}, we verify our method's edit consistency with quantitative experiments. 
%how consistent the transferred edits look to the human eye using different methods. 
%for some known attributes 

%We now turn to demonstrate that our method preserver the edits (quality), while helps save significant time and efforts for image editing process (efficiency).

% \vspace{-10pt}
% \paragraph{Baselines.} We consider two categories of methods: those that (i) perform point based image editing, and (ii) use text for image editing. We use DragGAN \cite{DragGAN}, UserControllableLT \cite{UserControllableLT}, and DragDiffusion \cite{DragDiffusion} as point-dragging baselines. For text-based methods, we compare with StyleCLIP \cite{StyleCLIP} and a popular state-of-the-art diffusion-based image editing method InstructPix2Pix \cite{InstructPix2Pix}. We use each method's official code.% for their evaluation.
%as a medium for that process

\vspace{-10pt}
\paragraph{Categories.} We evaluate on a variety of domains: Human faces (FFHQ) \cite{stylegan}, AFHQ Cats, Dogs \cite{afhq}, MetFaces \cite{ganwithlimiteddata}, Human bodies \cite{styleganhuman}. For each domain, we use the corresponding pretrained StyleGAN model to perform editing.  

\begin{figure*}[t]
\centering 
\includegraphics[width=0.86\textwidth]{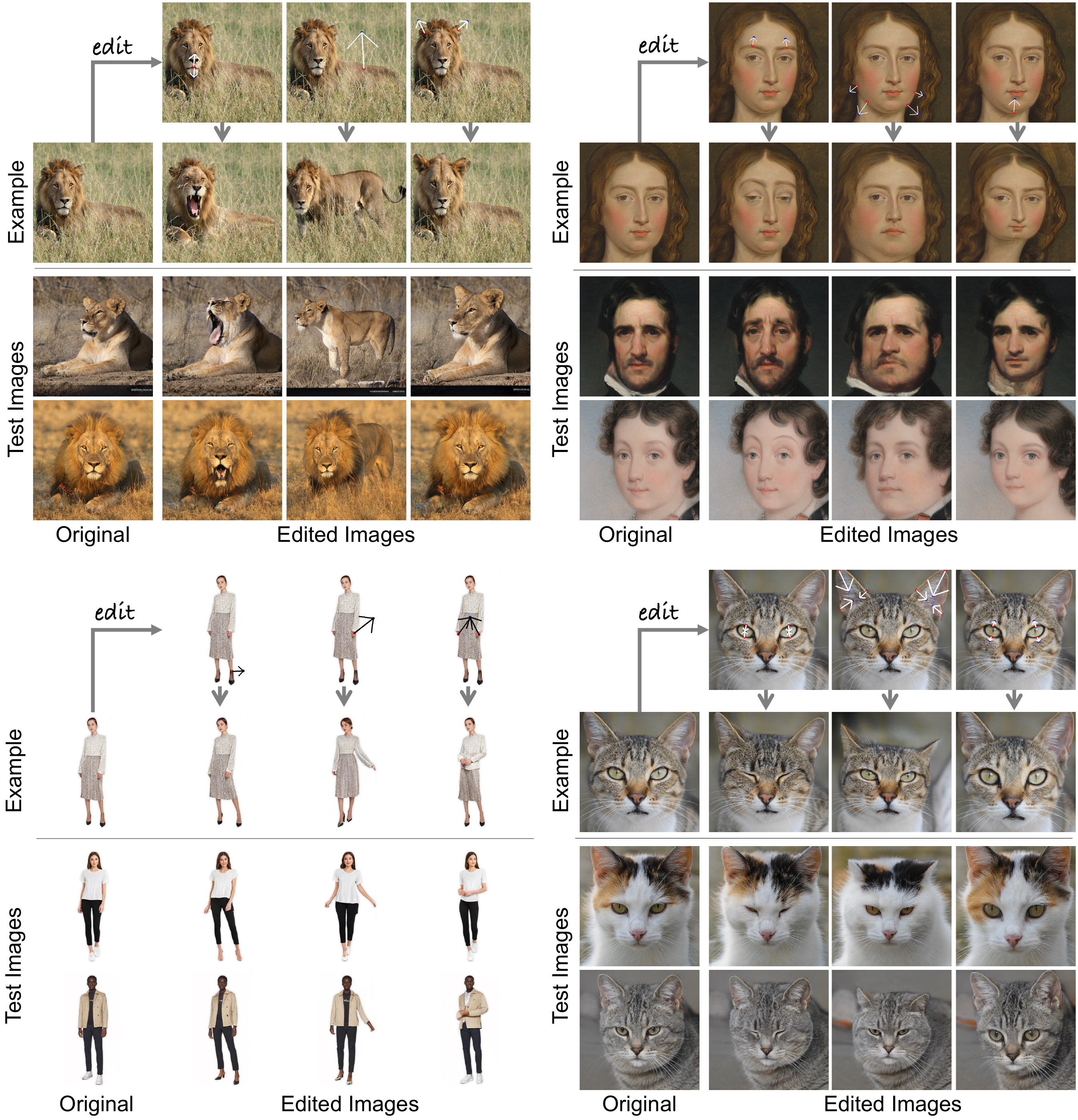}
% \fbox{\rule{0pt}{5in} \rule{.9\linewidth}{0pt}}
\vskip -0.1in
   \caption{Additional qualitative results on various domains.}
   \vspace{-3mm}
\label{fig:ours_other}
\end{figure*}

\subsection{Qualitative results}\label{sec:qual}

We start with qualitative comparisons with (i) interactive point-dragging, and (ii) text-based image editing baselines.
%image

%For the interactive point dragging setup, 
\vspace{-10pt}
\paragraph{Interactive point based editing.} We compare against DragGAN~\cite{DragGAN}, UserControllableLT~\cite{UserControllableLT}, and DragDiffusion~\cite{DragDiffusion}. For each baseline, we, as a user, desire to edit a bunch of images to have the same final state for an attribute. Fig.~\ref{fig:comparison-drag} shows the results of editing two kinds of edits to four images (leftmost column). In the left case, the goal is to make everyone smile by the same degree. In the right, it is to vertically compress everyone's face by the same degree. For all three baselines, we manually annotate points for each test image. For our method, we only annotate points and perform DragGAN-based edit on one image (top row), then automatically transfer the edit to the three other test images.

%First, we focus on the results from the baselines. 
For the smiling case (left), we notice that when DragGAN drags the upper/lower lips up/down respectively, it can either make people smile (1st, 2nd and 4th rows), or it can make people look shocked (blonde girl; 3rd row). Both edits are technically correct based on lips movement, but it won't match the user expectation of making everyone smile by the same degree. Since all four images are edited independently, this is not unexpected. The other two baselines sometimes have issues introducing the same edit across different images; e.g., UserControllableLT can make the blonde girl smile a little bit, but not the other images. In general, we notice in our experiments that to introduce the same exact edit that we want, we often need to play around with some hyperparameters (e.g., \# of iterations). In comparison, our method produces edited images that are smiling more equally (left) and have been compressed by a similar amount (right), and with much less human effort.  

%We can see that ours results is consistent with DragGAN, while does not require manually annotate test image again each time. Ours also show more consistency of the final state, especially in edit (e.g., frontal faces). Or in smiling faces, using annotate points can lead to undesired edits (e.g, open mouth, not smilling). Thus is exactly what we want to emphasize: Edit on single image can be tedious, or require human-supervision for each image to ensure the consistency between test images.

\vspace{-10pt}
\paragraph{Text-driven image editing.} We consider two baselines: StyleCLIP~\cite{StyleCLIP} and InstructPix2Pix~\cite{InstructPix2Pix}. Each takes an image and text as input to produce an edited image: $I \rightarrow I'$. Fig.~\ref{fig:comparison-text} shows the results of Ours vs. StyleCLIP (left) and Ours vs. InstructPix2Pix (right). (For the latter, we invert the output of InstructPix2Pix into StyleGAN's latent space using \cite{e4e} for our method.) In each case, for the baselines, we use the same text prompt, e.g., ``A photo of a person with pale skin'' for each edited image (in four rows). For ours, we capture the edit from the first editing result of a baseline (first row), and then automatically transfer the edit to the remaining three images. 
% and get the corresponding latent edit ($\Delta_w = w' - w$), and use it
%, where we compare our method to each baseline individually
%{t}
%to get the results from

Our goal with this particular setup is to test (i) how consistently the baselines introduce the edit denoted by text to different images (e.g., do all people become equally pale?), and (ii) how consistently our method captures and transfers the \emph{particular} edit of the first example to the rest of the images, irrespective of how good that first edit was. For (i), we find that StyleCLIP sometimes has issues; e.g., the \textit{paleness} of the edited face in first example is different from the second (first vs second rows). For InstructPix2Pix, sometimes it can give consistent results; the \textit{childness} of different faces seems similar. But, it can have consistency issues in other examples; e.g., the \emph{type/strength} of maleness introduced is different in each image. And this is where, we believe, the utility of our method lies: if one desires to edit every face to be \emph{male} in a particular way (i.e., to the same degree) as the first example, e.g., thick eyebrows \& light beard, our method has an advantage. %can do that to a decent degree.   

Finally, we show our method's results for non-facial domains in Fig.~\ref{fig:ours_other}. For the lion examples (top left), we see that it can preserve not just the type of edit, e.g., dragging to make the lion roar,  but also the strength of the roar. The strength being preserved can be observed more easily for the human body poses (bottom left), where we see that the extent of legs split, hand movements, is consistent enough to almost align the edited test images with the user edited one. Overall, results on these diverse set of domains highlight an observation that many kinds of edits can be thought of as a combination of \{type, strength\}, both of which can indeed be captured and transferred according to our needs.   

\begin{table}
\resizebox{ \columnwidth}{!}{%
% \begin{tabular}{lcccc>{\columncolor{yellow!20}}c}
\begin{tabular}{@{\extracolsep{4pt}}lccccc@{}}
\toprule
% & \multicolumn{2}{c}{Point Dragging} & \multicolumn{2}{c}{Time} & \multicolumn{2}{c}{Anno.}\\
& \multicolumn{2}{c}{Point Dragging} & \multicolumn{2}{c}{Time} & Anno.\\
\cline{ 2-3} \cline{4-5}
Method & Dist. & FID & 1 img & 1k imgs & \# imgs \\
\midrule
UserControl. \cite{UserControllableLT} & 13.64 & 25.32 & 0.03s & 30s & 1000 \\
DragGAN \cite{DragGAN} & 4.165 & 9.28 & 2s & 33.33m & 1000 \\
DragDiffusion \cite{DragDiffusion} & 26.56 & 36.55 & 60s & 16.67h & 1000 \\
% DragonDiffusion \cite{DragonDiffusion} & \textcolor{red}{18.51} & 0 & - & - \\
% \midrule
% StyleCLIP \cite{StyleCLIP} & - & - & 9.255 & 40.503\\
% SDEdit \cite{SDEdit} & - & - & 6.833 & 48.039 \\
% InstructP2P \cite{InstructPix2Pix} & - & - & 9.023 & 66.046 \\
% \midrule
% AdaTrans \cite{AdaTrans} & - & - & 0 & 0\\
Ours & 9.467 & 9.35 & 2s & 82s & 1\\
% Ours & 17.689 & 0 & 9.243 & 38.090\\
\bottomrule
\end{tabular}
} 
\caption{
Time is estimated for 1 point drag, without human annotation time. (82s includes 2s to perform edit on the example, 30s to optimize the editing direction, and 50s to transfer the edit to 1000 test images.) Our method requires only one image annotation in total, while the baselines need one annotation per test image.
}
\vspace{-0.09in}
\label{tab:quantitative}
\end{table}
\subsection{Quantitative results}\label{sec:quant}

%\ut{we wish to validate our observations from the previous section through a quantitative study}
Next, perform quantitative experiments following the setup proposed in~\cite{DragGAN,DragonDiffusion,DragDiffusion}. We first randomly sample 10 facial test images and pair them with a randomly sampled target face image. We use dlib-ml \cite{dlib} to detect keypoints in the test and target images. Our goal is to see if the keypoints in the test images can be moved to the target locations specified by keypoints in the target image. For the baselines \cite{UserControllableLT, DragGAN, DragDiffusion}, we perform dragging for each test image. For our method, we perform dragging on \emph{one} test image, and then transfer the edit to the remaining nine images. This is repeated 100 times; i.e., each time a random target image is paired with 10 random test images. 
%check this by 

We report the Euclidean distance between the keypoints of edited and target images in Table~\ref{tab:quantitative}. Even without requiring annotations for every test image, our method can move the points very close to the target; almost as close as DragGAN, which requires annotation each time. We also compute FID~\cite{fid} between the original test images and their edited versions to ensure that our method does not distort the image; and our quality is comparable to DragGAN. %Next, we study how long it takes to perform editing on so many images. \ut{DISCUSS}   

\section{Deeper Analysis}

We perform deeper analysis on the usefulness of our batch image editing framework by focusing on a specific edit: face pose rotation. In particular, if we wish to rotate many faces to front, how do we visualize the improvements brought by different components of our method? 
%, if any

\begin{figure}[t]
\centering 
\includegraphics[width=0.47\textwidth]{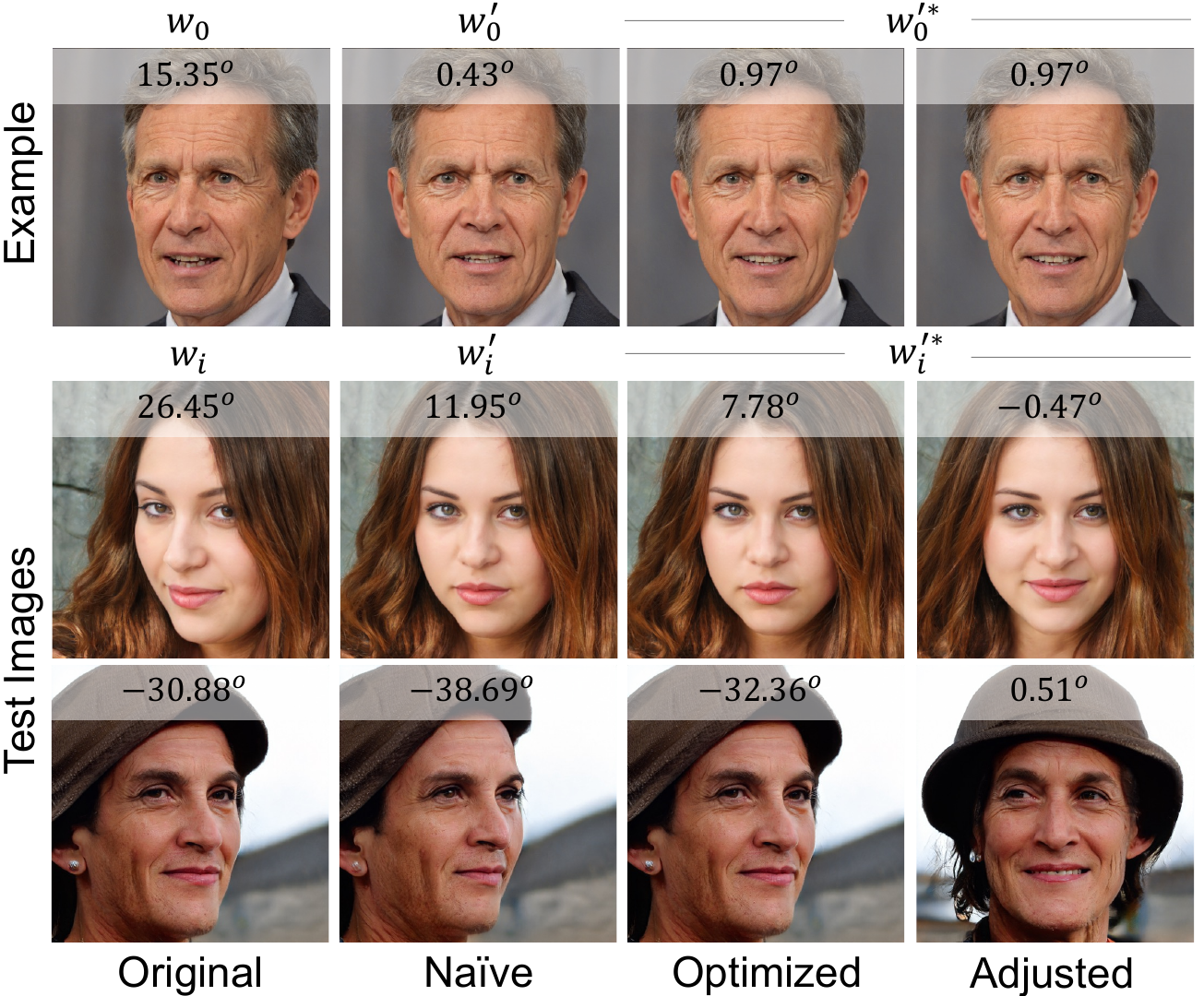}
% \fbox{\rule{0pt}{2in} \rule{.9\linewidth}{0pt}}
\vskip -0.1in
   \caption{Effect of adjusted and optimized editing for test images. Yaw degree of each image is provided on top of each images.}
   \vspace{-1mm}
\label{fig:ablation-qualitative}
\end{figure}

\begin{figure}[t]
\centering 
\includegraphics[width=0.47\textwidth]{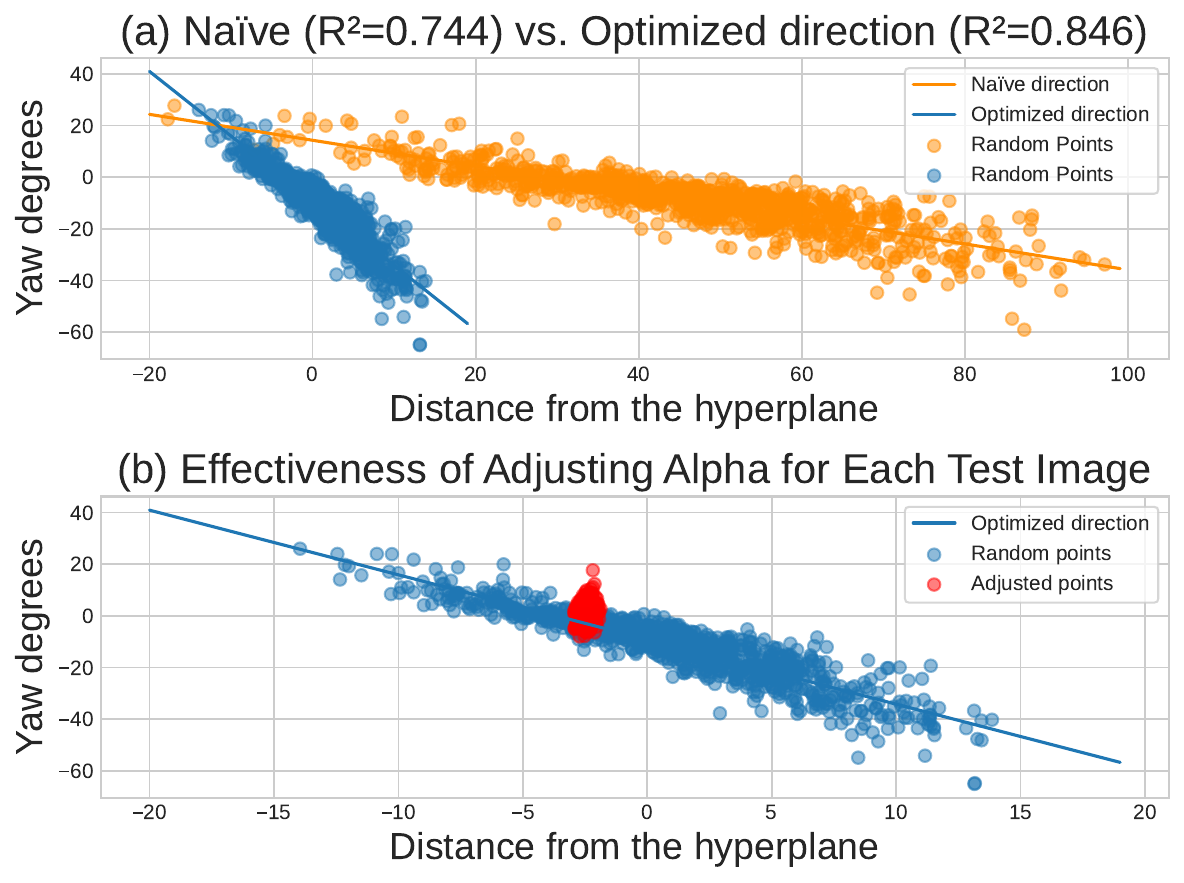}
% \includegraphics[width=0.49\textwidth]{figures/ablation-quantitative-swapped.pdf}
% \fbox{\rule{0pt}{2in} \rule{.9\linewidth}{0pt}}
\vskip -0.1in
   \caption{(a) After optimization, the editing direction is more linearly correlated with the yaw attribute. (b) With automatically adjusted editing scale for each test image.}
   \vspace{-3mm}
\label{fig:ablation-quantitative}
\end{figure}

Fig.~\ref{fig:ablation-qualitative} depicts the setup. We use DragGAN to perform dragging on one image (top; $w_{0} \rightarrow w'_{0}$). We then transfer the edit to 1000 other test images (two shown in rows 2-3). We see that \textit{naively} applying $\Delta_w = w'_{0} - w_{0}$ from DragGAN to test images cannot make them frontal (0\degree). Using our optimized $\Delta^{*}_w$ (without dynamic scaling) does help it bring closer to facing front. Does this mean that $\Delta^{*}_w$ might be a more \emph{global} direction? We study this in Fig.~\ref{fig:ablation-quantitative} (a), where we visualize the degree to which the correlation property of global directions (Eq.~\ref{eq:linear_correlation}) holds for $\Delta_w$ and $\Delta^{*}_w$ individually. We see that there is indeed a better correlation ($R^2$) between the distances of latents from hyperplanes and the Yaw degree of resulting images. (We predict Yaw using 6DRepNet~\cite{yawdetector}.)  However, as we discussed before, $\Delta^{*}_w$ in itself is insufficient: to completely bring the facial pose to front, we need to scale it with corresponding $\alpha$'s to bring them much closer to front (Fig.~\ref{fig:ablation-qualitative}, 4th column). To study this in a more systematic way, we visualize the effect on all 1000 test images. Similar to Fig.~\ref{fig:ablation-quantitative} (a), we visualize distance-to-hyperplane vs. Yaw degree for the original test images (blue) and edited images (red) in Fig.~\ref{fig:ablation-quantitative} (b). We can see that the variation in Yaw for the edited images \emph{collapses} at around 0\degree; i.e., they mostly face front as we would like.
%We can see that the variation in Yaw for the original test images somewhat \emph{collapses} at around 0\degree for the edited images; i.e., edited images mostly face front.    
%the facial pose using 
% (attributes)

We also compute the time it takes to perform editing in Table~\ref{tab:quantitative}, last three columns. Our method only requires annotating one image, while the baselines require annotating each image (e.g., 1000 total in this case). Note the times do not include human annotation time. 

%We conduct two main ablation study: (1) To show the effectivenes of each components, and (2) Show the advantages of batch-image editing vs single image editing.

%We consider the most intuitive problem setting: Rotate face to frontal. We will drag on one example, then trasfer to other test images.
%The example image is generated with random seed 123456. We randomly sample 1000 facial images to serve as test images.
% Problem: Frontal face. Example (random seed = 123456), Random 1000 images for test. First: Study ablation of each approachs. Second: Study the variant of DragGAN (transfer points)
%\paragraph{\textbf{Effects of optimizing and adjusting editing direction}}
%Using DragGAN's editing direction might not be globally consistent, as Fig.~\ref{fig:ablation-quantitative} (first subplot). After optimization, the $R^{2}$ score of optimized (0.846) is higher than the naive one (0.744).
%But while the optimized direction is more linear in nature, it does not turn every test points has yaw-degree to frontal (0).

%Using the adjusted formula Eq.~\ref{eq:adjusting_alpha} helps moving each point (blue) to the desired final state (frontal) (red).
\begin{table}
\resizebox{ \columnwidth}{!}{%
\begin{tabular}{lccc}
\toprule
& & \multicolumn{2}{c}{Time complexity}\\
\cline{3-4}
Method & MAE & Prepare & Inference\\
\midrule
Random & 11.295 $\pm$ 8.972 & - & -\\
DragGAN \cite{DragGAN} + GANgealing \cite{GANgealing} & 8.141 $\pm$ 7.221 & 30s & 2000s\\
Ours & 2.120 $\pm$ 1.818 & 32s & 50s\\
\bottomrule
\end{tabular}
}
\caption{
Ablation Study: DragGAN + GANgealing.
}
\vspace{-0.05in}
\label{tab:ablation-dragan}
\end{table}

\begin{figure}[t]
\centering 
\includegraphics[width=0.47\textwidth]{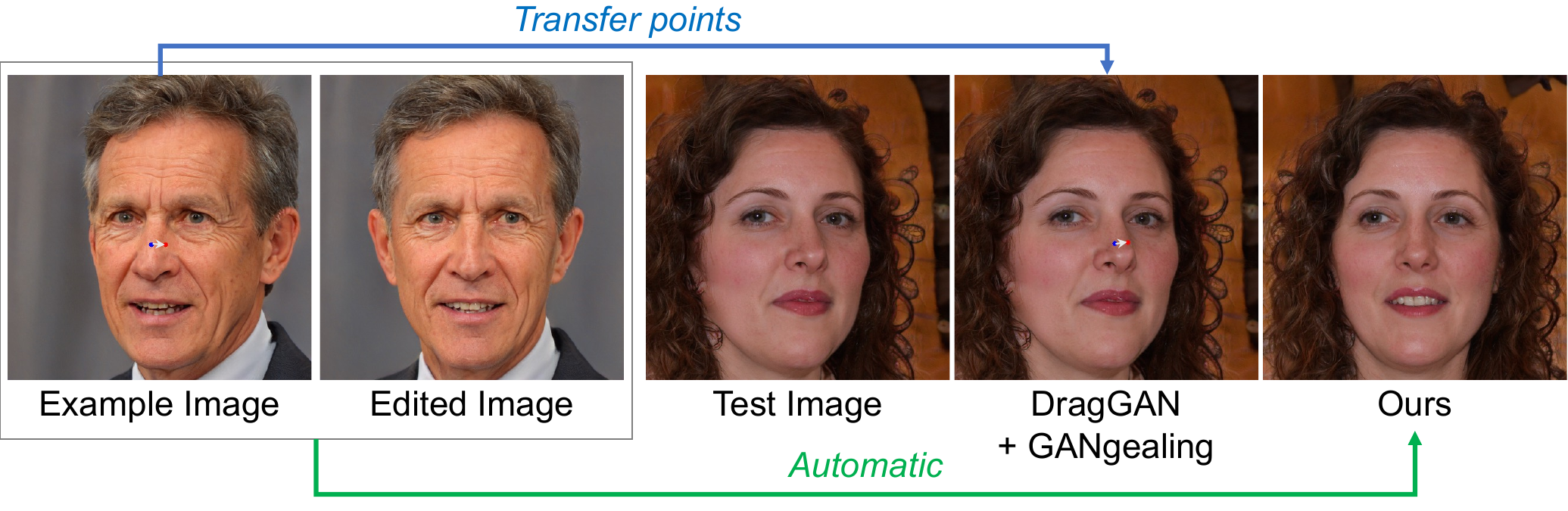}
% \fbox{\rule{0pt}{2in} \rule{.9\linewidth}{0pt}}
\vskip -0.12in
   \caption{Ablation study set up. We use GANgealing \cite{GANgealing} to transfer the annotated points from example to test image.}
   \vspace{-3mm}
\label{fig:ablation-set-up}
\end{figure}

\vspace{-10pt}
\paragraph{Other ways to automate batch image editing.} One baseline for batch image editing (i.e., so that all edited images achieve the same final state) is shown in Fig.~\ref{fig:ablation-set-up}. After the user annotates the source/target points in the example image, we use GANgealing~\cite{GANgealing} to transfer the points to corresponding locations in each new test image. The edited images are then obtained using DragGAN (DragGAN + GANgealing). We compare our method to this baseline when the user edits one face image to become front (Yaw=0\degree) and transferring it to 1000 other test images.

Results are shown in Table~\ref{tab:ablation-dragan}. We report mean absolute error (MAE) between the Yaw degree of the edited image and its ideal frontal image (Yaw=0\degree). `Random' shows the variations in Yaw for the original images (before editing). While `DragGAN + GANgealing' does help in reducing the variation (8.14 $<$ 11.29), our edited images are much more \textit{frontal}, with an MAE of 2.12. This is because the baseline has no way to bring all edited images to the same final state.  On top of that, since our method does not rely on additional information about keypoints (which `DragGAN + GANgealing' does), we believe that it is better suited for batch image editing for many kinds of domains (e.g., Fig.~\ref{fig:ours_other}), where we might not have such information.

\begin{figure*}
    \centering
\includegraphics[width=\textwidth,page=1]{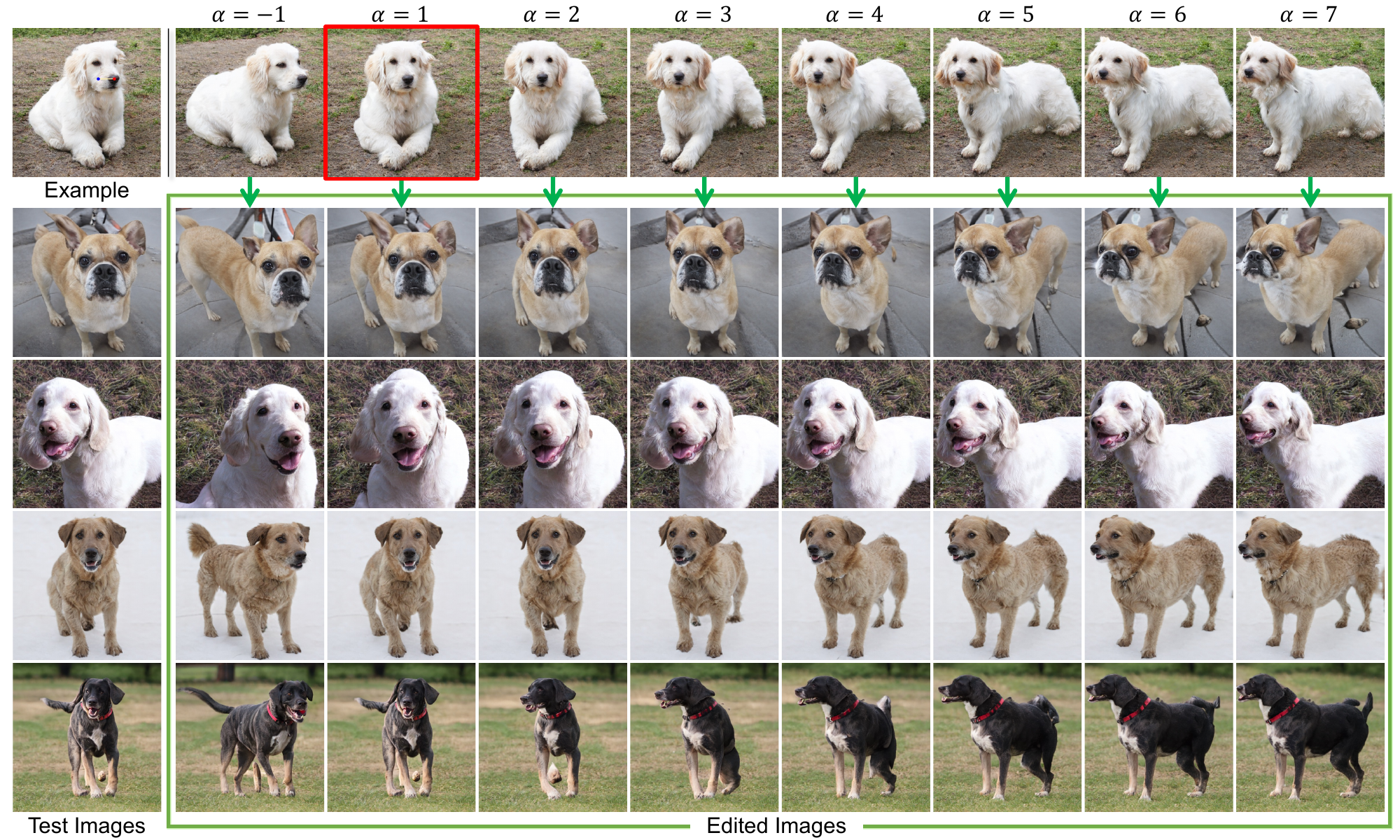}
    \caption{Interactive Batch Image Editing. As users adjust the editing strength in the example image, all test images will be automatically updated. (Red bounding boxes indicate the $\Delta_w$ according to the drag points).}
    \label{fig:interactive-dog}
\end{figure*}

\section{Conclusion}

We introduced the problem of interactive batch image editing.  Given a user edit in an example image, our approach automatically transfers that edit to other test images, maintaining a consistent final state of the edit across images. Extensive experiments demonstrated that our method produces comparable quality to state-of-the-art single-image-editing methods while saving significant time and human effort.  We are currently limited to StyleGAN models. Extending this problem and solution to diffusion-based models for more edits types would be an exciting future direction. %Another observation that we notice is that GANs space might not handle out-of-distribution well. At some extend, the edits will make images become distorted.

{
    \small
    \bibliographystyle{unsrtnat}
    \bibliography{main}
}

\section{Supplementary}\begin{figure*}
    \centering
\includegraphics[width=\textwidth,page=6]{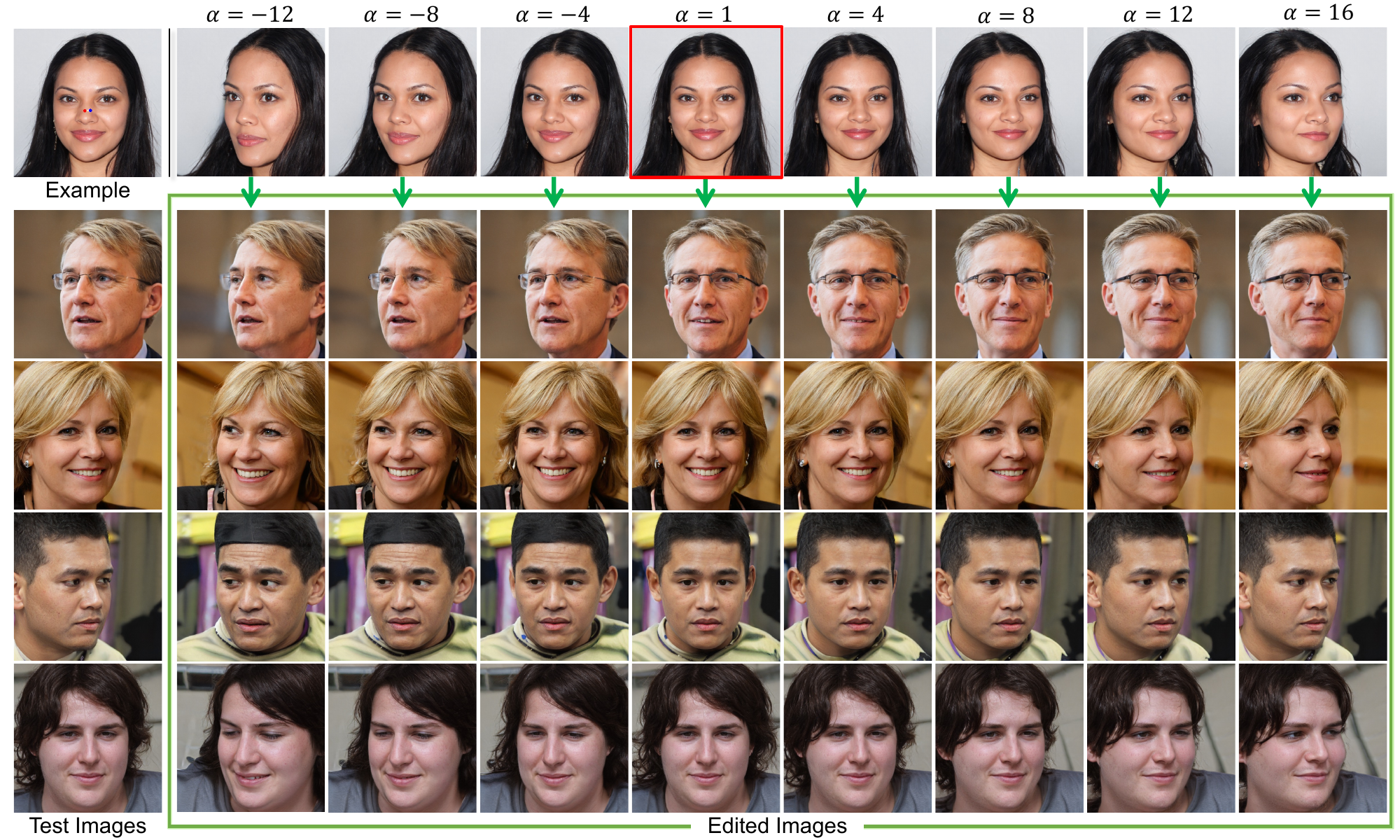}
    \caption{\textbf{Interactive Batch Image Editing}. As the user adjusts the editing strength $\alpha$ in the example image (first row), all test images will be automatically updated, ensuring consistency in the final state (yaw degree). (Red bounding box indicates the $\Delta_w$ yielded by dragging points).}
    \label{fig:interactive-pose}
    \vspace{-0.1in}
\end{figure*}
% \lht{do we use distributd training (the current description makes me think so)? if not, we just need to say that we train all models with a single rtx 3090. we can also move this to the last section as other experiments are more interesting}

% For StyleCLIP~\cite{StyleCLIP}, we user default 

% \paragraph{Content}
% This document provides additional qualitative results that could not be included in the main paper due to page limits.
We present results from Interactive Batch Image Editing in Section~\ref{sec:interactive}, where adjustments in the example are automatically updated to test images. Subsequently, we demonstrate the effectiveness of our method when multiple edits are applied to an example before reaching the final state in Section~\ref{sec:multiple-edits}. Limitations are discussed in Section~\ref{sec:limitation}, followed by additional qualitative results in Section~\ref{sec:additional-qualitative} and implementation details in Section~\ref{sec:implementation}.

\subsection{Interactive Batch Image Editing}
\label{sec:interactive}
\vspace{-0.1in}
We demonstrate the advantages of computing adaptive $\alpha$ values for individual example images.
Consider a scenario where the editing objective is to rotate and bring into frontal view a set of $n$ faces.
User will first annotate dragging points for example image to rotate this face to frontal (Fig.~\ref{fig:interactive-pose}, 1st row, $\alpha=1$).
Upon completion, each test images will also become frontal, each with unique editing strengths represented by ${\alpha_{1}, \alpha_{2}, ..., \alpha_{n}}$, computed using $\alpha_i = (w'_{0} - w_{i}) \cdot n$ with $n = \Delta^{*}_{w} / || \Delta^{*}_{w} ||$ (Fig.~\ref{fig:interactive-pose}, row 2-5, 5th column).

Now, if the user wishes to have the same faces facing slightly to the left, there is no need to re-annotate the original training examples and re-run the optimization process. In an interactive fashion, user can simply adjust the scaling of the $\alpha$ for the specific training example to achieve the desired edit (e.g., $\alpha=-4$), and all other $\alpha$ values will be automatically recalculated. Qualitative results are shown in Fig.~\ref{fig:interactive-pose}. As can be seen, all poses (yaw degrees) in test images are changing accordingly to the example image (yaw degree is consistent). It is worth noting that it takes roughly 0.05s to compute a new image (about 0.03s to recompute $\alpha$ for each test image and 0.02s to generate a new image). Thus, it is fast enough for real-time batch image editing.

\begin{figure*}[h]
    \centering
\includegraphics[width=0.95\textwidth,page=3]{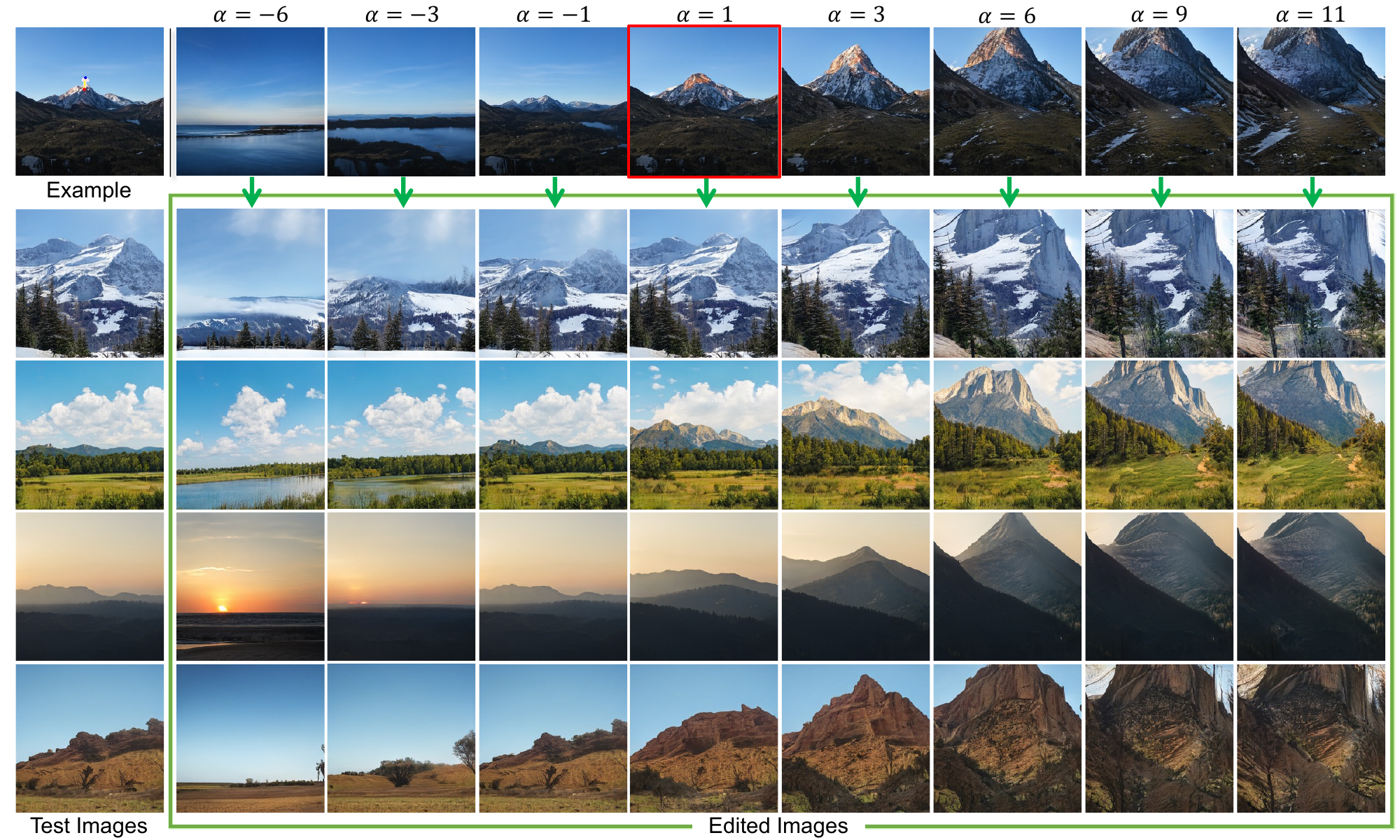}
    \caption{\textbf{Interactive Batch Image Editing.} As the user adjusts the editing strength $\alpha$ in the example image (first row), all test images will be automatically updated, ensuring consistency in the final state (mountain height). (Red bounding box indicates the $\Delta_w$ yielded by dragging points).}
    \vspace{-0.2in}
    \label{fig:interactive-mountain}
\end{figure*}

\begin{figure*}[h]
    \centering
\includegraphics[width=0.95\textwidth,page=5]{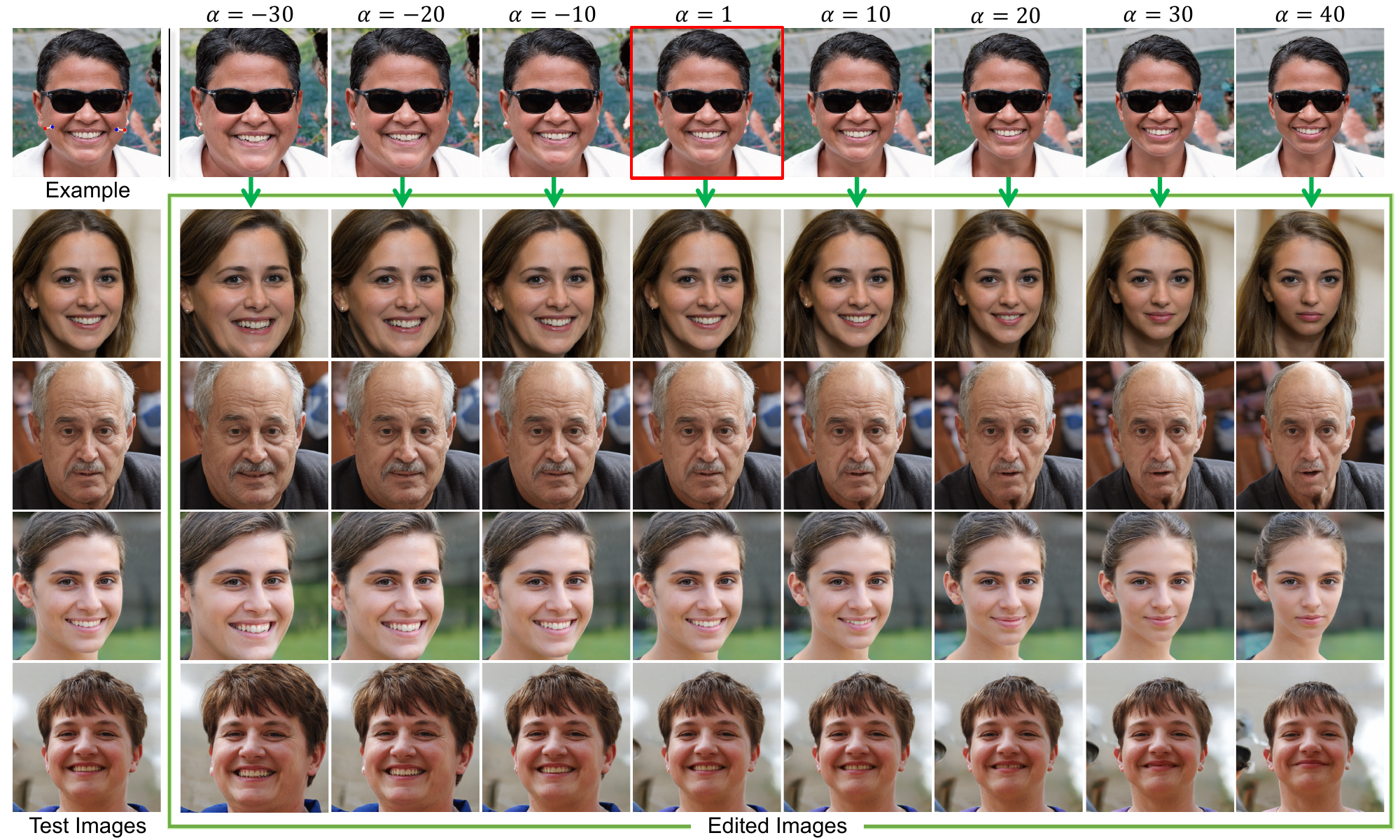}
    \caption{\textbf{Interactive Batch Image Editing.} As the user adjusts the editing strength $\alpha$ in the example image (first row), all test images will be automatically updated, ensuring consistency in the final state (facial shape). (Red bounding box indicates the $\Delta_w$ yielded by dragging points).}
    \vspace{-0.2in}
    \label{fig:interactive-face}
\end{figure*}

\begin{figure*}
    \centering
\includegraphics[width=0.95\textwidth,page=2]{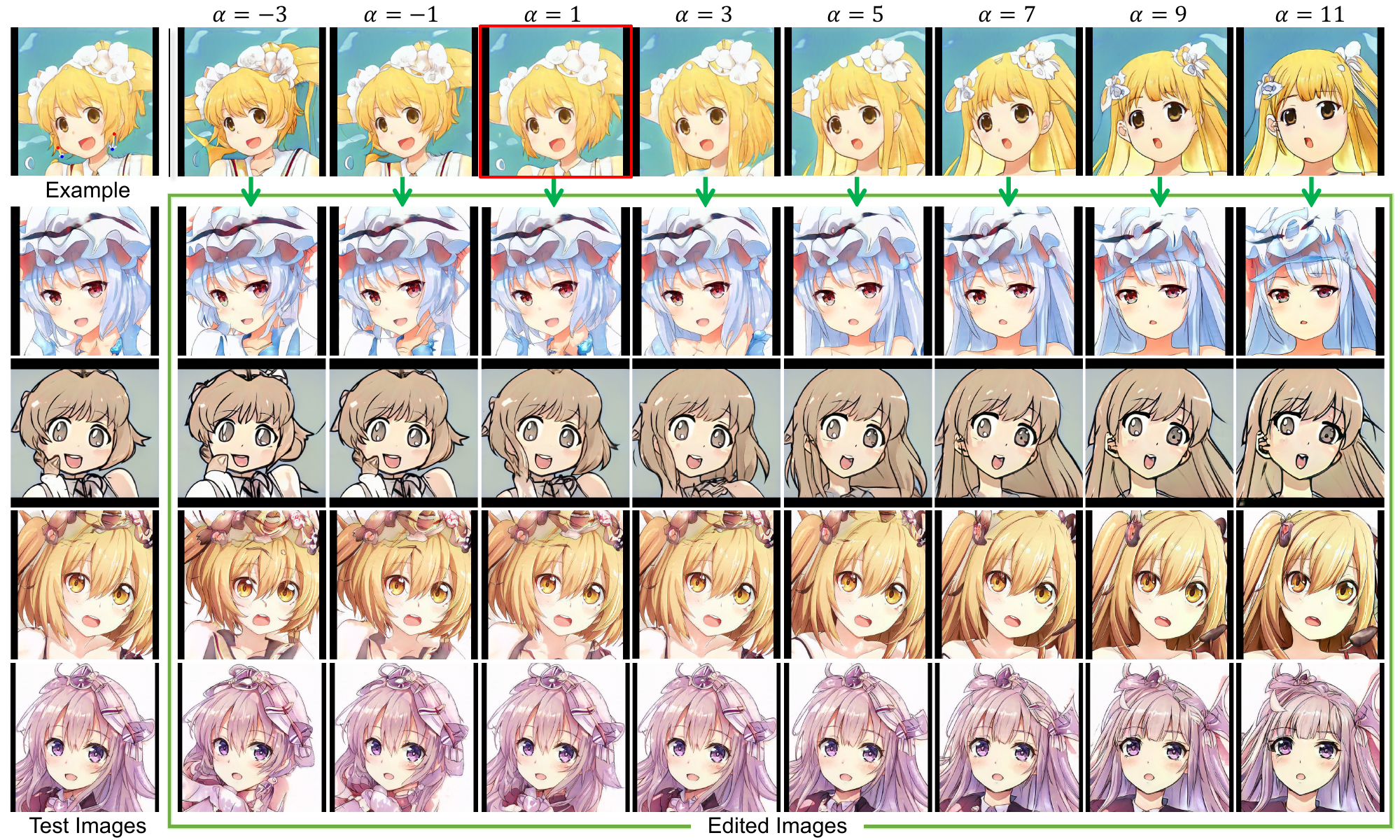}
    \caption{\textbf{Interactive Batch Image Editing}. As the user adjusts the editing strength $\alpha$ in the example image (first row), all test images will be automatically updated, ensuring consistency in the final state (hair length). (Red bounding box indicates the $\Delta_w$ yielded by dragging points).}
    \vspace{-0.2in}
    \label{fig:interactive-anime}
\end{figure*}

\begin{figure*}
    \centering
\includegraphics[width=0.95\textwidth,page=7]{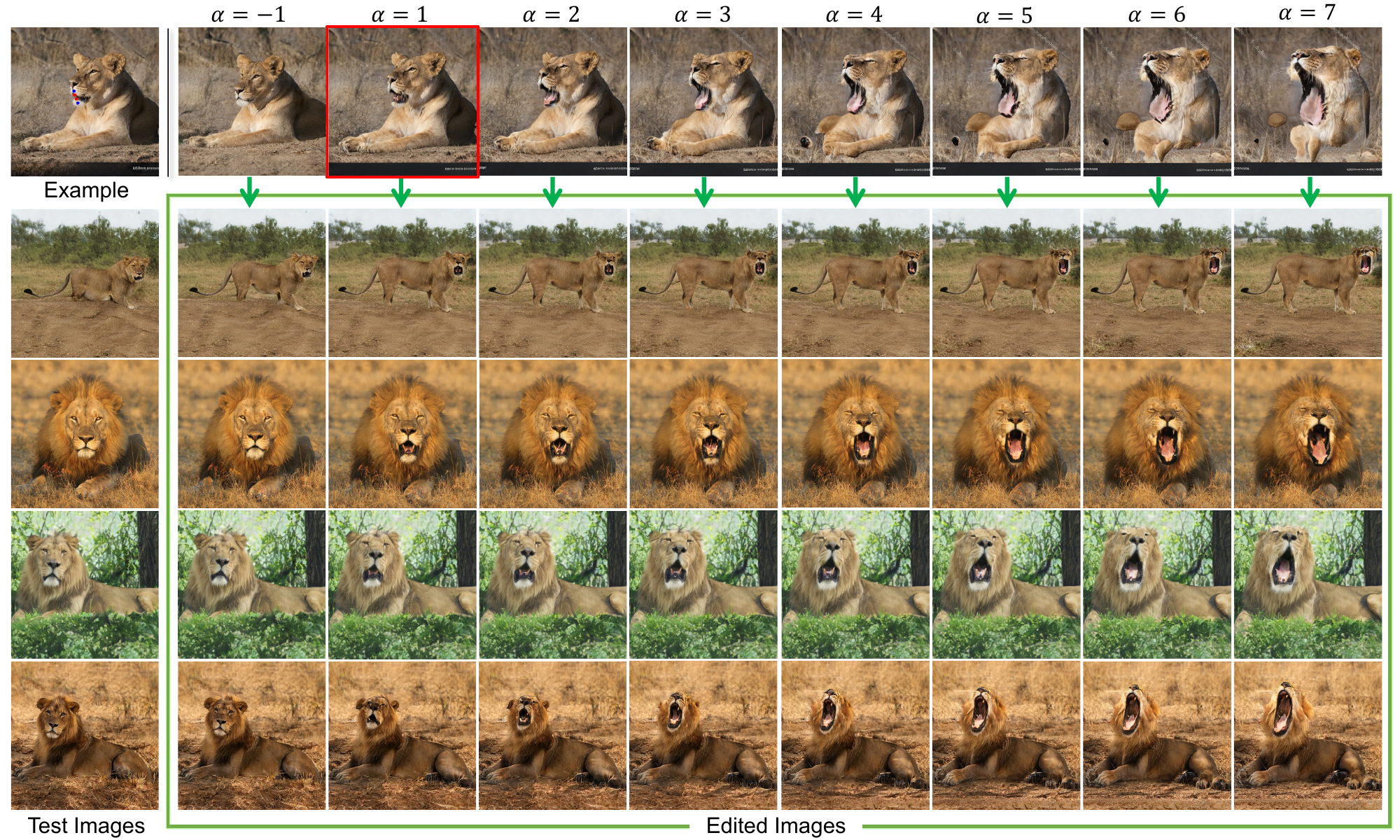}
    \caption{\textbf{Interactive Batch Image Editing}. As the user adjusts the editing strength $\alpha$ in the example image (first row), all test images will be automatically updated, ensuring consistency in the final state (roar degree). (Red bounding box indicates the $\Delta_w$ yielded by dragging points).}
    \vspace{-0.2in}
    \label{fig:interactive-tiger}
\end{figure*}

\begin{figure*}
    \centering
\includegraphics[width=0.95\textwidth,page=4]{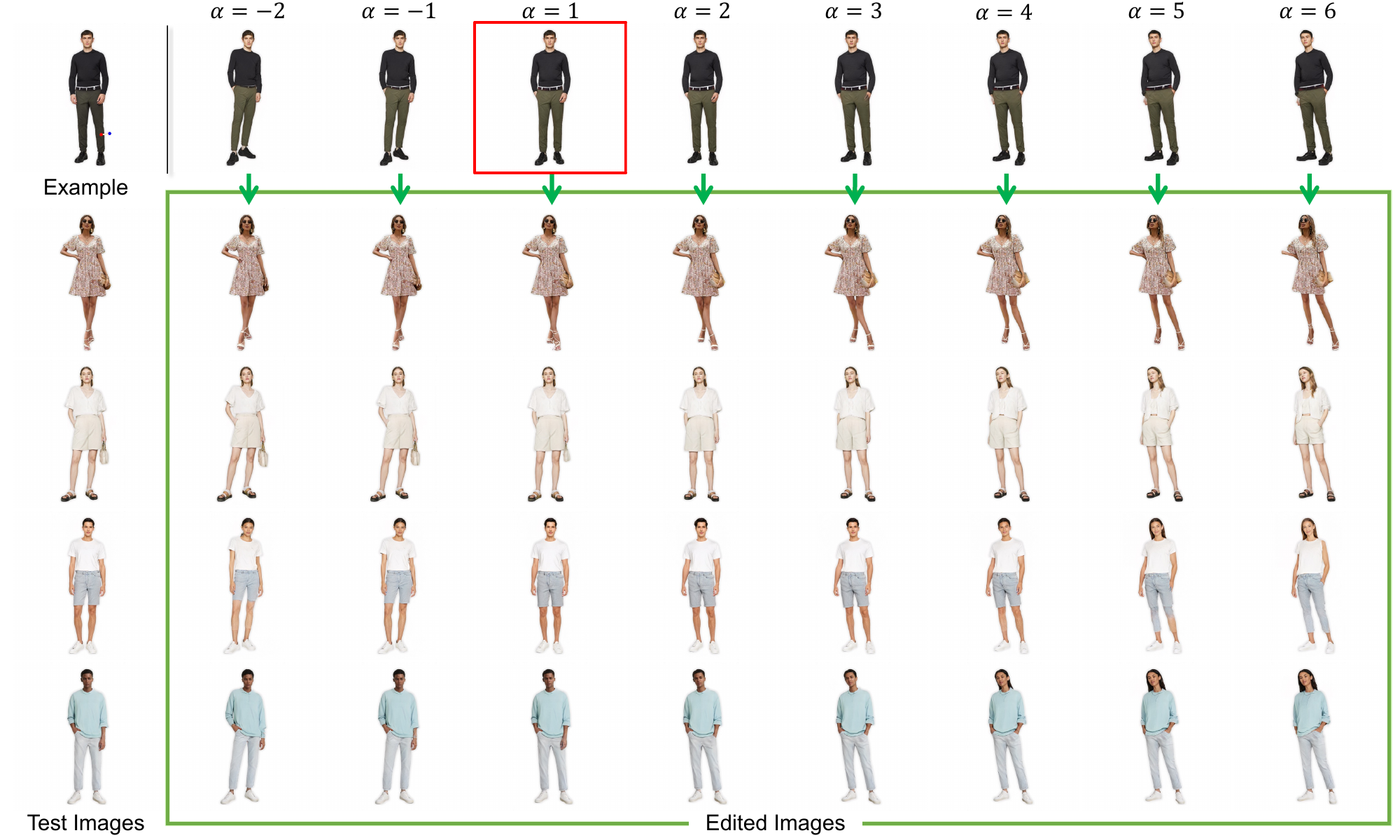}
    \caption{\textbf{Interactive Batch Image Editing}. As the user adjusts the editing strength $\alpha$ in the example image (first row), all test images will be automatically updated, ensuring consistency in the final state (leg position). (Red bounding box indicates the $\Delta_w$ yielded by dragging points).}
    \vspace{-0.2in}
    \label{fig:interactive-human}
\end{figure*}

Additional qualitative results for interactive batch image editing are provided in Fig.~\ref{fig:interactive-dog} (dog pose rotation), Fig.~\ref{fig:interactive-mountain} (mountain enlargement/removal), Fig.~\ref{fig:interactive-face} (face slimming/enlargement), Fig.~\ref{fig:interactive-anime} (anime hair shortening/lengthening),  Fig.~\ref{fig:interactive-tiger} (tiger roaring) and Fig.~\ref{fig:interactive-human} (model leg pose).

\begin{figure*}
    \centering
\includegraphics[width=\textwidth]{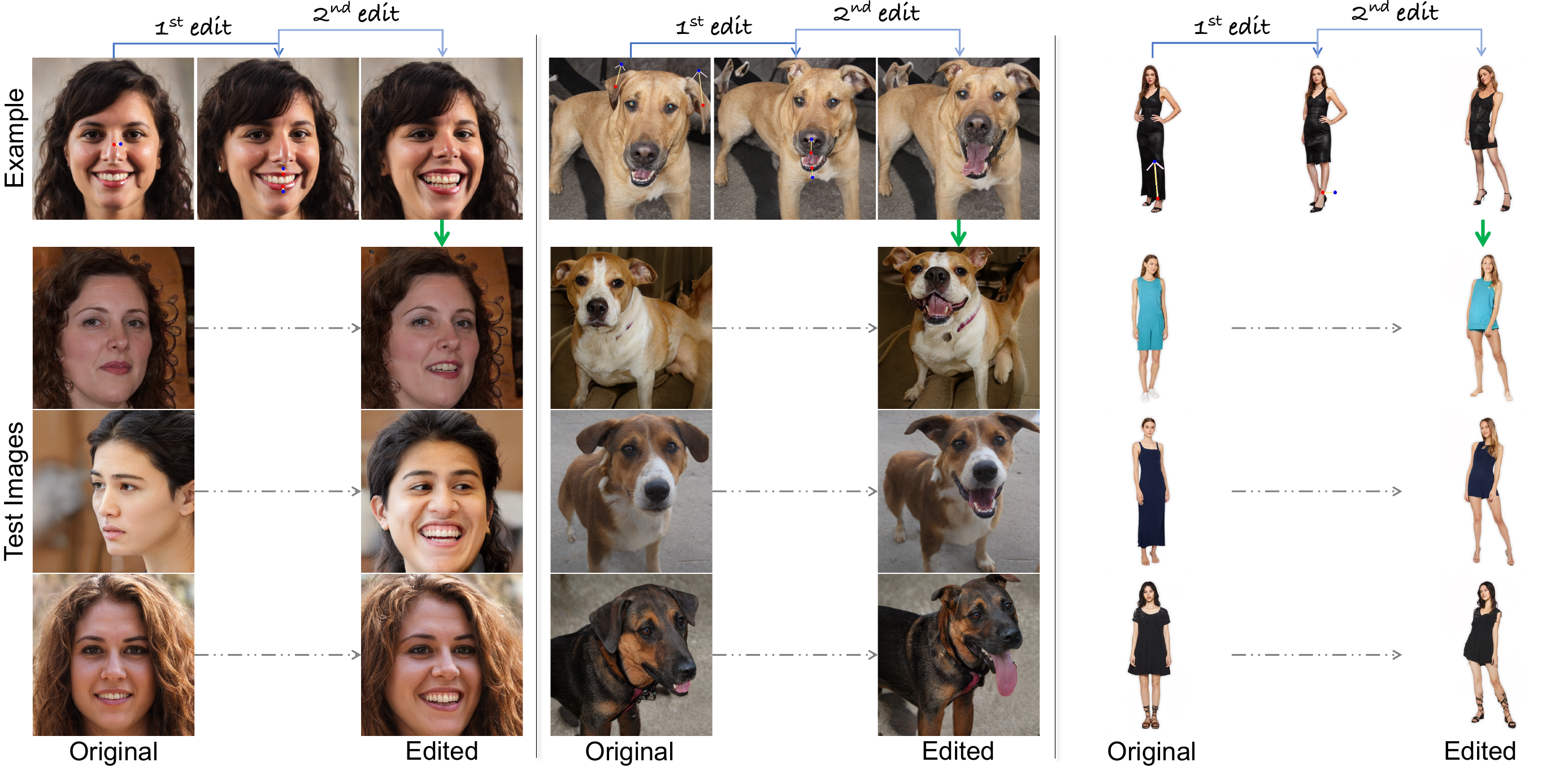}
    \caption{\textbf{Multiple edits} can be applied to example image before being transferred to test images.}
    \label{fig:mutiple-edits}
\end{figure*}

\subsection{Multiple Edits}
\label{sec:multiple-edits}
Multiple edits can be applied to the example to reach the final state, which is then transferred to the test images without intermediate steps. For example, a user may choose to initially rotate the face to the left (1st edit) and subsequently enhance the person's smile (2nd edit). The resulting final state incorporates both edits. Our method can directly transfer the final state to test images, including both edits (face rotation and smiling) (Fig.~\ref{fig:mutiple-edits}, columns 1-3).

Two additional examples of multiple edits are presented in Fig.~\ref{fig:mutiple-edits}. Columns 4-6 showcase uplifting the dog's ear (1st edit) and then opening the mouth (2nd edit). Columns 7-9 illustrate shortening the dress (1st edit) and adjusting the pose (2nd edit). Despite variations in the initial states of the test images (e.g., variations in mouth openness), our methods ensure consistency in the final states across test images (e.g., all dogs have an open mouth and uplifted ears).

\begin{figure*}[h]
    \centering
\includegraphics[width=\textwidth]{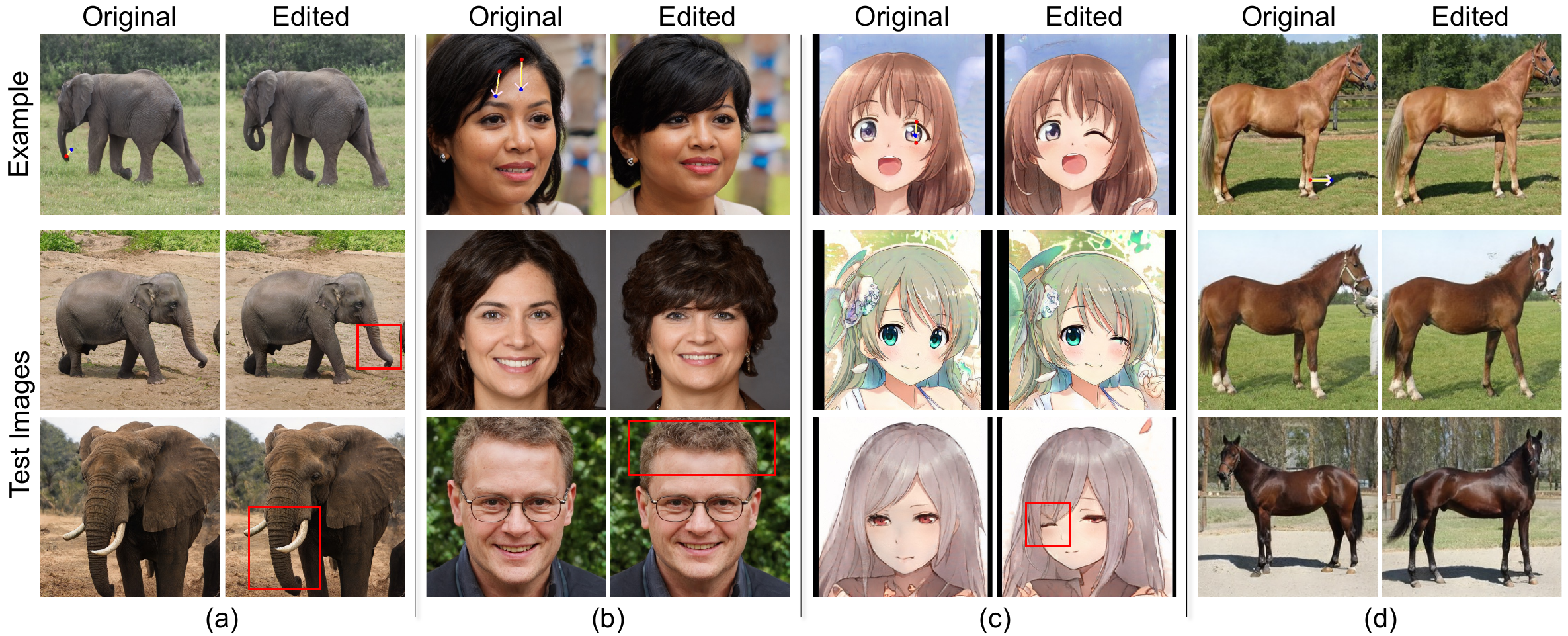}
    \caption{\textbf{Limitations.} (a) Failure Case: Our method may encounter challenges in capturing fine details (e.g., curling trunk of an elephant). (b) Example-Test Similarity: For optimal results, the example and test images should belong to the same semantic domain (e.g., both featuring long hair) to ensure correctly transferred edits. (c-d) Interesting Cases: Edits can be mistakenly interpreted, resulting in unexpected outcomes such as winking in the wrong eye (c) or unintentionally flipping the horse (d).}

    \label{fig:limitation}
\end{figure*}

\subsection{Limitations}
\label{sec:limitation}
% We notice that our method might fail to transfer small details (e.g., curling elephant trunk Fig.~\ref{fig:limitation}a).
% When the test images is not semantically similar with the example (e.g., test image has short hair, while example has long hair), the transferred results is not good as well (Fig.~\ref{fig:limitation}b).

% Sometimes we find the edits might be mistakes (e.g., make a person wink at left-eye, but when transferred, sometimes it transferred to the right eyes Fig.~\ref{fig:limitation}c). Another interesting cases is speading the horse legs, sometimes it flip the horse to same position as the example iamge as well (Fig.~\ref{fig:limitation}d).
While our method has demonstrated effectiveness in various applications, there are some limitations/failure cases that we notice. (1) Failure to capture small details: Our method may encounter challenges in accurately transferring small details (e.g., curling elephant trunk, Fig.~\ref{fig:limitation}a). (2) Semantic dissimilarity in example/test images: When test images deviate significantly in semantic content from the example (e.g., vastly differing in hair lengths), the transferred results may exhibit suboptimal performance (Fig.~\ref{fig:limitation}b). (3) Potential editing errors: In certain scenarios, the editing might be misinterpreted and lead to unexpected outcomes. For example, attempts to make a person wink at the left eye may occasionally lead to the wink being transferred to the right eye (Fig.~\ref{fig:limitation}c). Another example is pose changes, when adjusting the pose of horse legs, there are instances where the outcome unexpectedly mirrors the horse in the same position as the example image (Fig.~\ref{fig:limitation}d).
\subsection{Additional Qualitative Results}
\label{sec:additional-qualitative}
\begin{figure*}
    \centering
\includegraphics[width=0.85\textwidth,page=1]{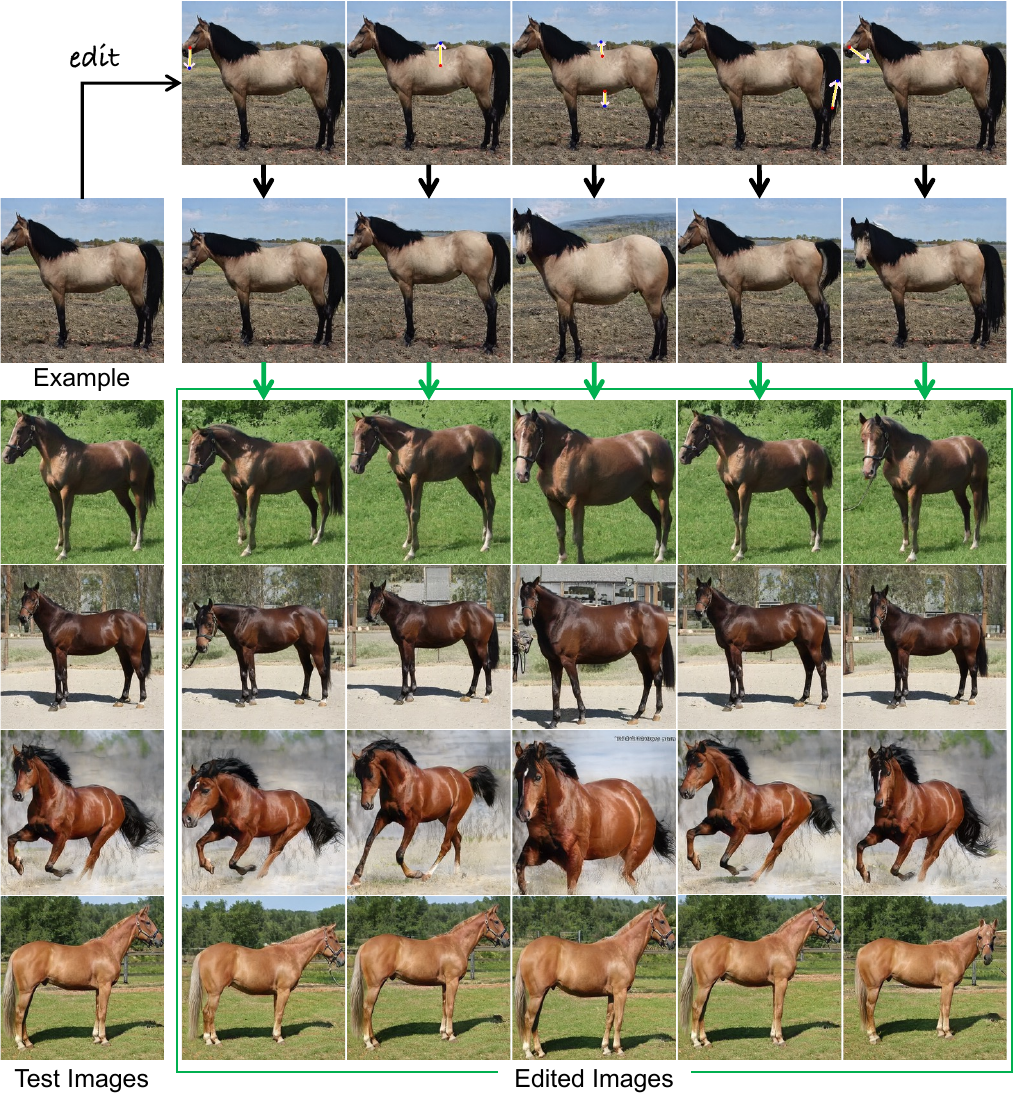}
    \caption{Additional qualitative results on Horses.}
    \label{fig:extra-domain1}
\end{figure*}

\begin{figure*}
    \centering
\includegraphics[width=0.85\textwidth,page=2]{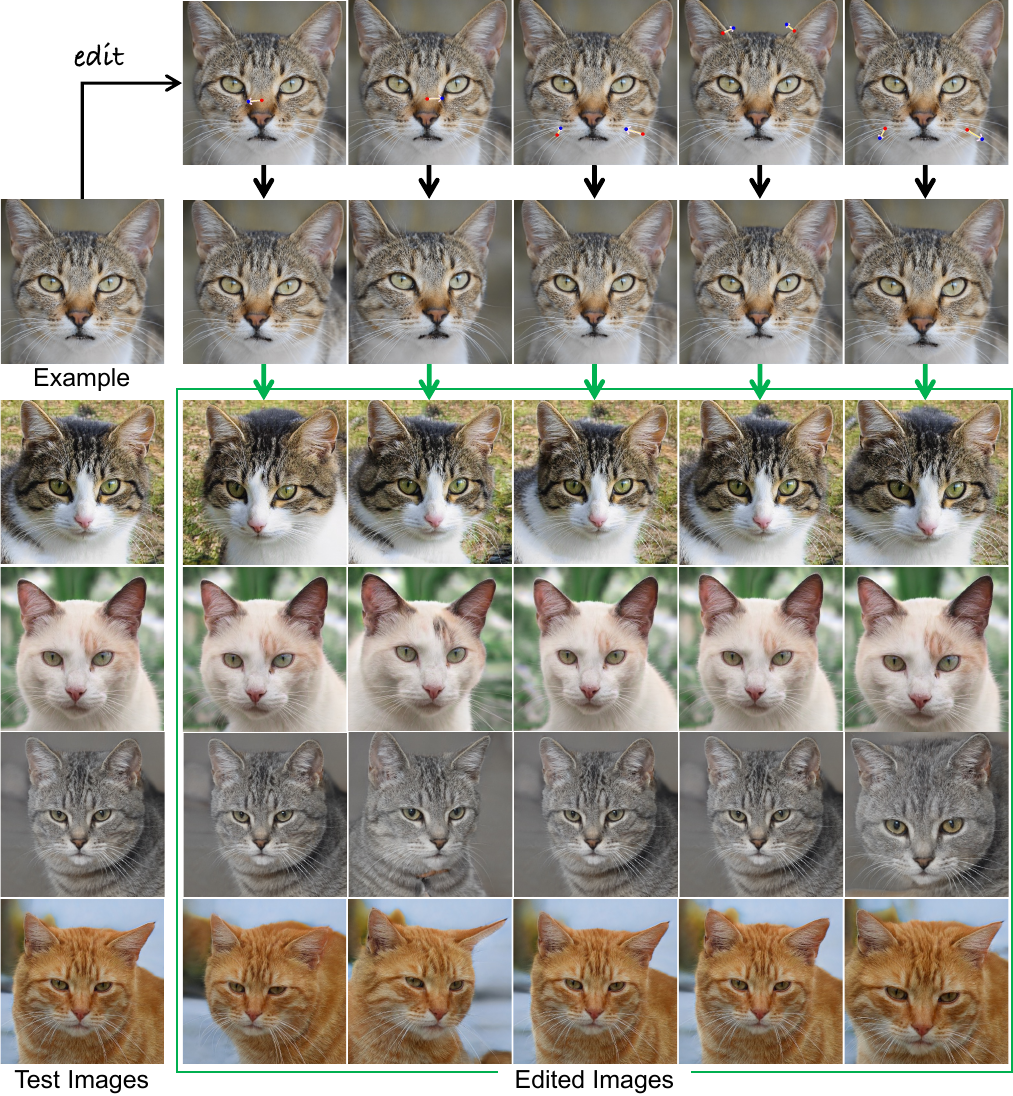}
    \caption{Additional qualitative results on Cats.}
    \label{fig:extra-domain2}
\end{figure*}

Along with the domains presented in the main paper (Human faces (FFHQ) \cite{stylegan}, Lions, Dogs \cite{afhq}, MetFaces \cite{ganwithlimiteddata}, Human bodies \cite{styleganhuman}), we provide additional qualitative results for Cats~\cite{afhq}, and Horses~\cite{lsun} domains in Fig.~\ref{fig:extra-domain2},~\ref{fig:extra-domain1} respectively.

% \begin{figure*}
%     \centering
% \includegraphics[width=0.85\textwidth,page=3]{figures/supp-extra domain.pdf}
%     \caption{Additional qualitative results on Cars.}
%     \label{fig:extra-domain3}
% \end{figure*}

\subsection{Implementation Details}
\label{sec:implementation}
We use the AdamW optimizer~\cite{adamw} to optimize $\Delta^*_w$ for 1000 iterations, with a learning rate of $0.001$ and a batch size of 16. All experiments are performed on a single NVIDIA RTX 3090 machine.

% {
%     \small
%     % \bibliographystyle{ieeenat_fullname}
%     \bibliographystyle{unsrtnat}
%     \bibliography{main}
% }
% \end{document}

% \pagebreak
% \clearpage


\end{document}